\documentclass[aps,pre,twocolumn,superscriptaddress,floatfix]{revtex4-2}
\usepackage{amsmath}
\usepackage{amssymb}
\usepackage{amsthm}
\usepackage{dcolumn}
\usepackage{mdwlist}
\usepackage{bm}
\usepackage{graphicx}
\usepackage{algorithm}
\usepackage{algpseudocode}
\usepackage[table]{xcolor}
\usepackage{subfig}
\usepackage{calc}
\usepackage{xcolor}
\usepackage{hyperref}
\usepackage[capitalise]{cleveref}
\usepackage{xurl}

\begin{document}

\title{Topological Learning in Multi-Class Data Sets}

\author{Christopher Griffin}
\affiliation{
	Applied Research Laboratory,
	The Pennsylvania State University,
    University Park, PA 16802
    }
    
\author{Trevor Karn}
\affiliation{
	School of Mathematics,
	University of Minnesota,
	Minneapolis, MN 55455
	}
    
\author{Benjamin Apple}
\affiliation{
	Naval Surface Warfare Center
	Carderock MD, 20817
    }

\date{\today~-~Preprint}

\begin{abstract} We specialize techniques from topological data analysis to the problem of characterizing the topological complexity (as defined in the body of the paper) of a multi-class data set. As a by-product, a topological classifier is defined that uses an open sub-covering of the data set. This sub-covering can be used to construct a simplicial complex whose topological features (e.g., Betti numbers) provide information about the classification problem. We use these topological constructs to study the impact of topological complexity on learning in feedforward deep neural networks (DNNs). We hypothesize that topological complexity is negatively correlated with the ability of a fully connected feedforward deep neural network to learn to classify data correctly. We evaluate our topological classification algorithm on multiple constructed and open-source data sets. We also validate our hypothesis regarding the relationship between topological complexity and learning in DNN's on multiple data sets.
\end{abstract}

\maketitle

\section{Introduction}

The use of deep learning methods in particular and artificial intelligence in general has become ubiquitous in science (see \cite{RBWV20,CHN18,G20,LZLS20,MCG20,SYZS18,WAU20,HSN20,IMWD20,AGS21,LT21,AT22,CM17,RLPE22,SHJ21,WASM21,SH20,DGGM21,KPN22,KHLA20} for a small example set). Recent work has shown that DNN's can generalize fundamental physical principles like symmetry \cite{LT21}. Yet, we still lack a fundamental understanding of when these techniques can be successfully applied \cite{W96,WM97,B22,SCHC22} and in some sense the \textit{no free lunch} theorems \cite{W96,WM97} imply it is impossible to know \textit{a priori} whether a data set is amenable to an off-the-shelf deep learning approach. Despite the fact that deep learning methods seem to frequently work, we may simply be observing a positive result bias \cite{J09}. A recent article by Das Sarma proposes that knowledge of when AI/ML techniques will fail will be crucial to their continued application in physics \cite{S23}. 

In this paper, we develop an approach for building topological information on multi-class data. This leads to the creation of a topologically inspired algorithm for classifying data, which is fully interpretable. We compare results from this algorithm to well-known off-the-shelf classifiers, including deep learning classifiers. As a consequence, we develop numerical tools to study the hypothesis that deep learning methods are susceptible to failure on classification problems when the underlying topology of the data is complex. We define (topological) complexity in terms of the topological information provided by the data itself. Thus, we propose a hypothesis based on data topology to explain why certain data sets are amenable to deep learning methods and to test when a classification problem may be amenable to deep learning.  

We validate our approach using multiple publicly available data sets as well as mathematically constructed data sets, helping to validate our underlying hypothesis. This paper focuses on classification \cite{KS17} problems. We do not consider more general approximation problems \cite{EPGB21} that can be solved with DNN's. Extending results from this work to more general approximation problems is left for future work. This work is part of a larger effort by mathematicians to understand neural networks using topological methods \cite{HS17,GL22}. 

This work is complementary to the work by Naitzat, Zhitnikov and Lim \cite{NZL20} who  study the impact of neural network layers on the topological structure of data in classification problems. Their approach uses persistent homology, whereas we draw inspiration from computational topology but define a specific topological structure that respects class information. Results in this paper are also related to work on the performance of deep learning as interpreted using differential topology and manifold learning. This is studied extensively by Buchanan, Gilboa and Wright \cite{BGW20} and Cohen et al. \cite{CCLS20}. 

The remainder of this paper is organized as follows: In \cref{sec:TopModel,sec:ClassAlg} we discuss the computational topology algorithms used in this paper. We provide detailed experimental results on topological classification for multiple data sets in \cref{sec:TopClassRes}. We study the problem of learning in the Math Dice Jr. and show that training failures are correlated with the topological complexity of the underlying data set in \cref{sec:MathDiceJr}. We then validate the hypotheses set forth in \cref{sec:MathDiceJr} using a secondary data set in \cref{sec:Carpet}. Conclusions are presented in \cref{sec:Conclusions}.

\section{Topological Features for of Multi-Class Data}
\label{sec:TopModel}
Consider a multi-class data set $(X_1,\dots,X_N)$, where $X_i \subseteq \mathbb{R}^n$ for $i=1,\dots,N$. Here, the classes range from $1$ to $N$. A classifier is a mapping $C:\mathbb{R}^n \to \{1,\dots,N\}$ that (correctly) assigns an arbitrary point $\mathbf{x}$ to one of the classes, assuming that the $N$ classes fully partition $\mathbb{R}^n$. The problem of ``learning'' such a classifier has been exhaustively studied (see e.g., \cite{HSD00}). We now consider this problem from a topological perspective and develop a method for extracting relevant topological features. Our approach is inspired by topological data analysis \cite{G07,C09}, which provides methods for extracting topological information about a topological space $\mathcal{T}$ on which a data set $X \subseteq \mathbb{R}^n$ resides.   

Consider a set $\mathcal{T} \subseteq \mathbb{R}^n$ as a topological space with a metric $d(\cdot,\cdot)$. We will be using the metric topology by default. Assume this space is partitioned into subspaces $\mathcal{T}_1,\dots,\mathcal{T}_N$ so that 
\begin{equation*}
    \mathcal{T} = \bigcup_i \mathcal{T}_i.
\end{equation*}
We assume that each data set $X_i$ consists of points drawn from the subspace $\mathcal{T}_i$. A point $\mathbf{x}$ is on the boundary of $\mathcal{T}_i$ and $\mathcal{T}_j$ if for all $\epsilon$ the ball of radius $\epsilon$ centered at $\mathbf{x}$ denoted $B_\epsilon(\mathbf{x})$ has non-empty intersection with both $\mathcal{T}_i$ and $\mathcal{T}_j$. We hypothesize that ``complex'' boundaries imply a harder classifier learning problem, and now proceed to formalize what we mean by this intuitive statement. We make use of constructs from both point-set and algebraic topology. See \cite{M18,M19} for complete details on these subjects. 

For the remainder of this paper, we assume a covering of $\mathcal{T}_i$ is a collection of points and radii, $C_i = \{(\mathbf{c}_{i_1},r_{i_1}),\dots,(\mathbf{c}_{i_{n_i},r_{i_{n_i}}})\}$ so that for all $\mathbf{x} \in \mathcal{T}_i$ there is a $j \in \{i_1,\dots,i_{n_i}\}$ such that $\mathbf{x} \in B_{r_{i_j}}(\mathbf{c}_{i_j})$, though this can be generalized to arbitrary sets rather than balls. By a sub-covering of $C_i$, we mean a subset of the covering that also acts as a covering of $\mathcal{T}_i$.

We can use the data to construct an approximate covering that respects class boundaries by solving the following (simple) optimization problems,
\begin{equation*}
\forall \mathbf{x}_{i_j} \in X_i\left\{
\begin{aligned}
\min \;\; & {r_{i_j}}\\
s.t. \;\; & d(\mathbf{x}_{i_j},\mathbf{y}) \leq r_{i_j} \quad \forall \mathbf{y} \in \bigcup_{j \neq i} X_j
\end{aligned}\right..
\end{equation*}
This finds the distance to the closest data point with class different from the class of the point $\mathbf{x}_{i_j}$. Then an approximate covering for $\mathcal{T}_i$ is given by, 

\begin{equation*}
    C_i = \left\{(\mathbf{x}_{i_1}, r_{i_1}), \dots, (\mathbf{x}_{i_{n_i}}, r_{i_{n_i}})\right\}.
\end{equation*}

When $X_i$ is large, this may not be a computationally efficient cover because of its size. To find a smaller sub-cover, define a directed graph $\vec{G}(C_i)$ with vertex set $X_i$ and edge $E[\vec{G}(C_i)]$ defined so that,
\begin{equation*}
    (x_{i_j}, x_{i_k}) \in E[\vec{G}(C_i)] \iff 
    d(x_{i_j}, x_{i_k}) < r_{i_j}.
\end{equation*}
That is, an edge points from point $x_{i_j}$ to $x_{i_k}$ if the ball centered at $x_{i_j}$ covers $x_{i_k}$. To construct the sub-cover, we build a minimal dominating set \cite{BM08} for $\vec{G}(C_i)$. That is, a set of vertices so that every vertex in $\vec{G}(C_i)$ is either in this set or covered by (adjacent to) an element in this set. It is known that finding such a set is $\mathrm{NP}$-hard \cite{BM08}, however a minimal dominating set can be approximated using the greedy algorithm shown in \cref{alg:AMSC}.
\begin{algorithm}[H]
  \caption{Approximate Minimal Sub-Cover}
  \label{alg:AMSC}
   \begin{algorithmic}[1]
   \State Set $G_\text{now} = \vec{G}(C_i)$.
   \While{$G_\text{now}$ has at least one vertex}
   \State Add the vertex $v^*$ with the largest out-degree in $G_\text{now}$ and its corresponding radius to the dominating set $C_i^*$.
   \State Remove $v^*$ and its neighbors from $G_\text{now}$.
   \EndWhile
   \end{algorithmic}
\end{algorithm}
The resulting covering $C_i^*$ approximates a minimum sub-cover of $C_i$. This is the algorithm implemented in our experiments. We also note that \cref{alg:AMSC} can be replaced with a version that uses only radius information and is useful when constructing the graph $\vec{G}(C_i)$ is computationally intractable (see \cref{app:AltAlg}). 

One of the main problems of algebraic topology is the classification of spaces in terms of the number of holes or voids present in the space \cite{M18}. Homology theory provides an approach to computing these properties by transforming an arbitrary space into a topologically equivalent simplicial complex \cite{M18}. A simplicial complex can be understood in the context of a hypergraph on a set of vertices. A hypergraph $H = (V,E)$ is a set of vertices $V$ along with a set of hyper-edges $E$, where if $e \in E$, then $e \subseteq V$. Hyper-edges, unlike ordinary edges, can have any cardinality up to the number of vertices in the hyper-graph. A hypergraph is a simplicial complex if its edge set has the property that it is closed under the operation of taking subsets. That is, if $e$ is a hyper-edge, then any subset $f \subset e$ is also a hyper-edge. Let $H$ be a simplicial complex. The skeleton (or $1$-skeleton) of $H$ is the graph constructed from the vertex set of $H$ and the cardinality two edges of $H$; i.e., the usual graph-theoretic edges made of pairs of vertices. Complete details are given in \cite{M18}. Once a simplicial complex is constructed for a topological space, numerical linear algebra can be used to construct a Betti sequence $\vec{\beta} = (\beta_0,\beta_1,\dots)$, which provides relevant topological information. Each entry in the sequence is a non-negative integer that counts the number of holes (voids) of a given dimension present in the space. In particular, $\beta_0$ counts the number of components, $\beta_1$ counts the number of holes (insides of circles), $\beta_2$ counts the number of voids (insides of hollow spheres) etc. 

To construct a simplicial complex $H_i$ representing $X_i$ (and hence $\mathcal{T}_i$), we define a graph $G_i = (C^*_i, E_i)$ using the points in $C_i^*$ as the vertices. The graph $G_i$ will serve as the $1$-skeleton of $H_i$. From the topological data analysis perspective, the points in $C_i^*$ are ``witness points''. Given a data set $X \subseteq \mathbb{R}^n$, a witness set is a (small)  set $W \subset X$ that can be used to construct a simplicial complex that correctly represents the topological features in the data set $X$, i.e., the topological features of the space $\mathcal{T}$ in which the data set $X$ resides.

The edge set of $G_i$ is given by the edge rule,
\begin{equation}
    \{x_{i_j},x_{i_k}\} \in E_i \iff B_{r_{i_j}}(x_{i_j}) \cap B_{r_{i_k}}(x_{i_k}) \neq \emptyset.
    \label{eqn:Cech}
\end{equation}
That is, a simple edge is present if and only if the balls centered at the points $x_{i_j}$ and $x_{i_k}$ in the sub-cover intersect. The graph $G_i$ is the $1$-skeleton of the \v{C}ech complex $\text{\v{C}}(C_i^*)$, in which a hyper-edge is present if and only if the balls of the vertices occurring in the hyper-edge have non-empty intersection. For the purposes of this paper, we will not use the \v{C}ech complex, but we define $H_i$ to be the clique complex $\mathrm{Cl}(C^*_i)$, where $\{x_{i_{k_1}},\dots, x_{i_{k_m}}\} \in \mathrm{Cl}(C^*_i)$ if and only if $\{x_{i_{k_1}},\dots, x_{i_{k_m}}\}$ is a clique (or subgraph of a clique) in $G_i$. Here, a clique in a graph is a complete subgraph that is itself not contained in a larger complete subgraph \cite{G23}. We make this choice for $H_i$ for computational expediency. In general, the clique complex will have fewer topological features (necessarily) than the \v{C}ech complex and will differ primarily in small-scale topological features. As such,  the clique complex seems to represent features at a scale relevant to the classification problem. It follows by the nerve lemma \cite{EH10} that the topological features of the various spaces $\mathcal{T}_i$ ($i=1,\dots,N$) should be preserved at the scale of the classes if the cover is sufficiently dense. 

\section{Classification with the Topological Cover}
\label{sec:ClassAlg}
Given the multi-class data set $(X_1,\dots,X_N)$, let $(\mathcal{B}_1,\dots,\mathcal{B}_N)$ be the collections of balls generated by the covers, $C_1^*,\dots,C_N^*$ built using the approach described in the previous section. That is,
\begin{equation*}
    \mathcal{B}_i = \bigcup_j B_{r_{i_j}}(x_{i_j}),
\end{equation*}
is the set of balls covering the set $X_i$ and determined from the minimum cover creation process. The set $\mathcal{B}_i$ acts as an approximation to the topological space on which the data in $X_i$ lie. 

Suppose $\mathbf{x}$ is an unclassified point. We can classify $\mathbf{x}$ by testing whether $\mathbf{x} \in \mathcal{B}_i$ for each $i \in \{1,\dots,N\}$. If there is exactly one $i$ for which this is true, then this is the class assigned to $\mathbf{x}$. If $\mathbf{x} \in \mathcal{B}_i$ is true for no $i$, then we compute the distance,
\begin{equation*}
    d(\mathbf{x},\mathcal{B}_i) = \min_j d\left[\mathbf{x}, B_{r_{i_j}}(x_{i_j})\right],
\end{equation*}
where $d\left[\mathbf{x},B_{r_{i_j}}(x_{i_j})\right]$ is the point-to-set distance from $\mathbf{x}$ to the ball $B_{r_{i_j}}(x_{i_j})$ that is induced from the natural metric. We then assign $\mathbf{x}$ as,
\begin{equation*}
C(\mathbf{x}) = \arg\min_i d(\mathbf{x},\mathcal{B}_i).
\end{equation*}
If $\mathbf{x} \in \mathcal{B}_i$ is true for multiple $i$, then we use a nearest neighbors approach, computing,
\begin{equation*}
    \tilde{d}(\mathbf{x},\mathcal{B}_i) = \min_j d\left[\mathbf{x}, x_{i_j}\right]. 
\end{equation*}
Here we use the centers of the balls covering $X_i$, rather than the balls themselves, since the point is already covered by at least one ball. We then assign the class to $\mathbf{x}$ as 
\begin{equation*}
C(\mathbf{x}) = \arg\min_i \tilde{d}(\mathbf{x},\mathcal{B}_i).
\end{equation*}

The entire process is summarized in \cref{alg:TopClassify}.
\begin{algorithm}[H]
  \caption{Topological Classification}
  \label{alg:TopClassify}
   \begin{algorithmic}[1]
   \State Compute the set
   \begin{equation*}
       I(\mathbf{x}) = \left\{i \in \{1, \dots, N\} : \mathbf{x} \in \mathcal{B}_i\right\}.
   \end{equation*}
   \If{$|I(\mathbf{x})|=1$}
   \State Assign $C(\mathbf{x})$ the unique element of $I(\mathbf{x})$.
   \EndIf
   \If{$|I(\mathbf{x})| = 0$}
   \State $C(\mathbf{x}) = \arg\min_i d(\mathbf{x},\mathcal{B}_i)$.
   \EndIf

   \If{$|I(\mathbf{x})| > 1$}
   \State $C(\mathbf{x}) = \arg\min_i \tilde{d}(\mathbf{x},\mathcal{B}_i)$.
   \EndIf

   \end{algorithmic}
\end{algorithm}

\section{Results on Topological Classification}
\label{sec:TopClassRes}
We illustrate the topological covering and classification algorithms on several different data sets. We compare the topological classification results to deep neural network classifiers and random forests (where appropriate), which are de facto standards for classification. In our experiments, we used  standard feedforward neural networks with a ramp (ReLU) activation function between the layers and a softmax (Boltzmann distribution) as the final output layer. We describe neural network structures using a tuple of layer sizes. By way of example, the neural network structure $(8,4,2)$ has a linear layer of dimension 8 with ramp activation followed by a linear layer of dimension 4 followed by a ramp followed by a linear layer of dimension 2 followed by a two-class softmax classifier. All neural networks were implemented in Mathematica 13 using the built-in neural network tools. All neural network and random forest training used the default (automatic) settings in Mathematica.

\subsection{Complex Boundaries in Two Dimensions}\label{sec:SineWave}
\begin{figure}[htbp]
\centering
\includegraphics[width=0.8\columnwidth]{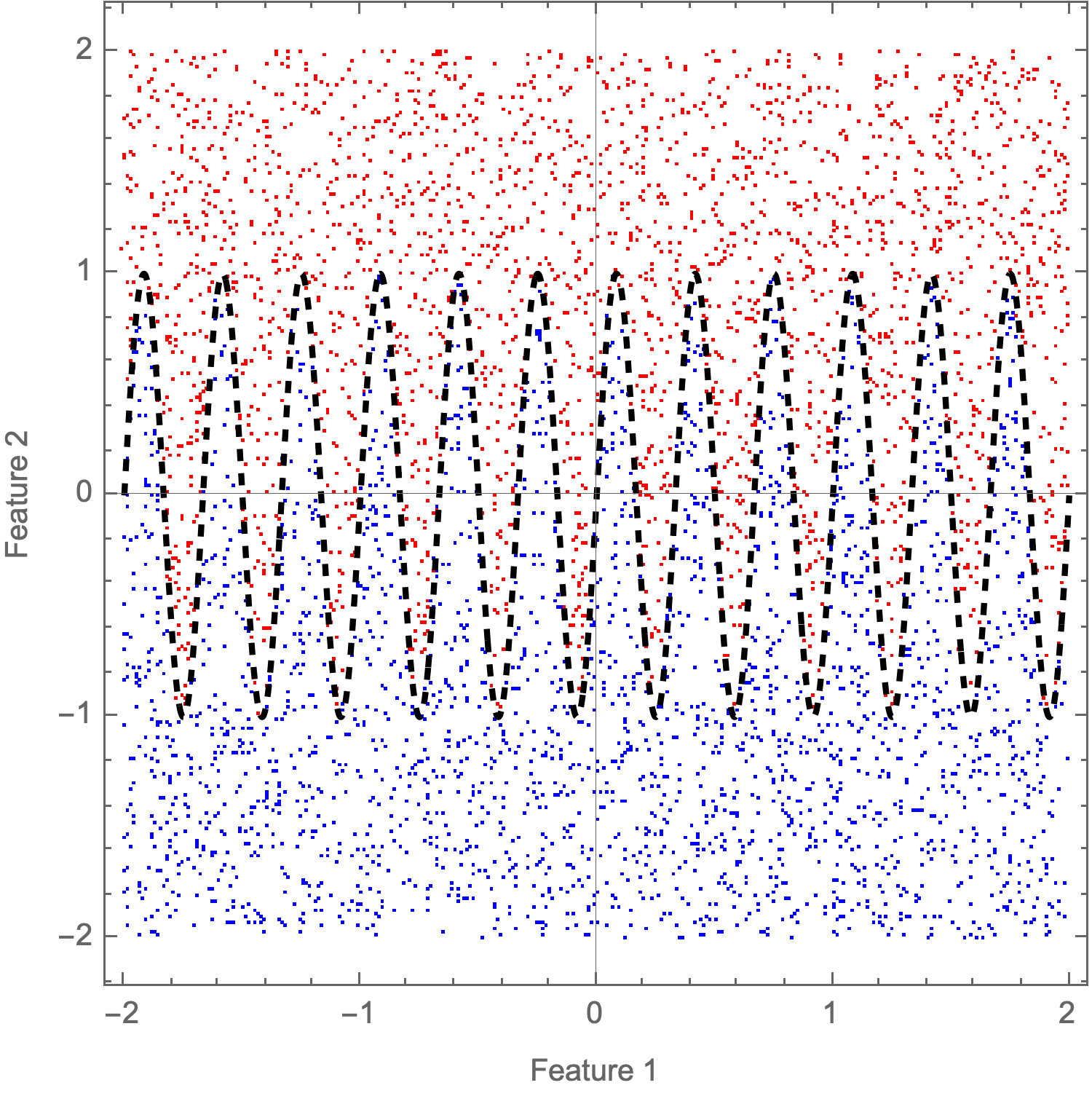}\\
\includegraphics[width=0.8\columnwidth]{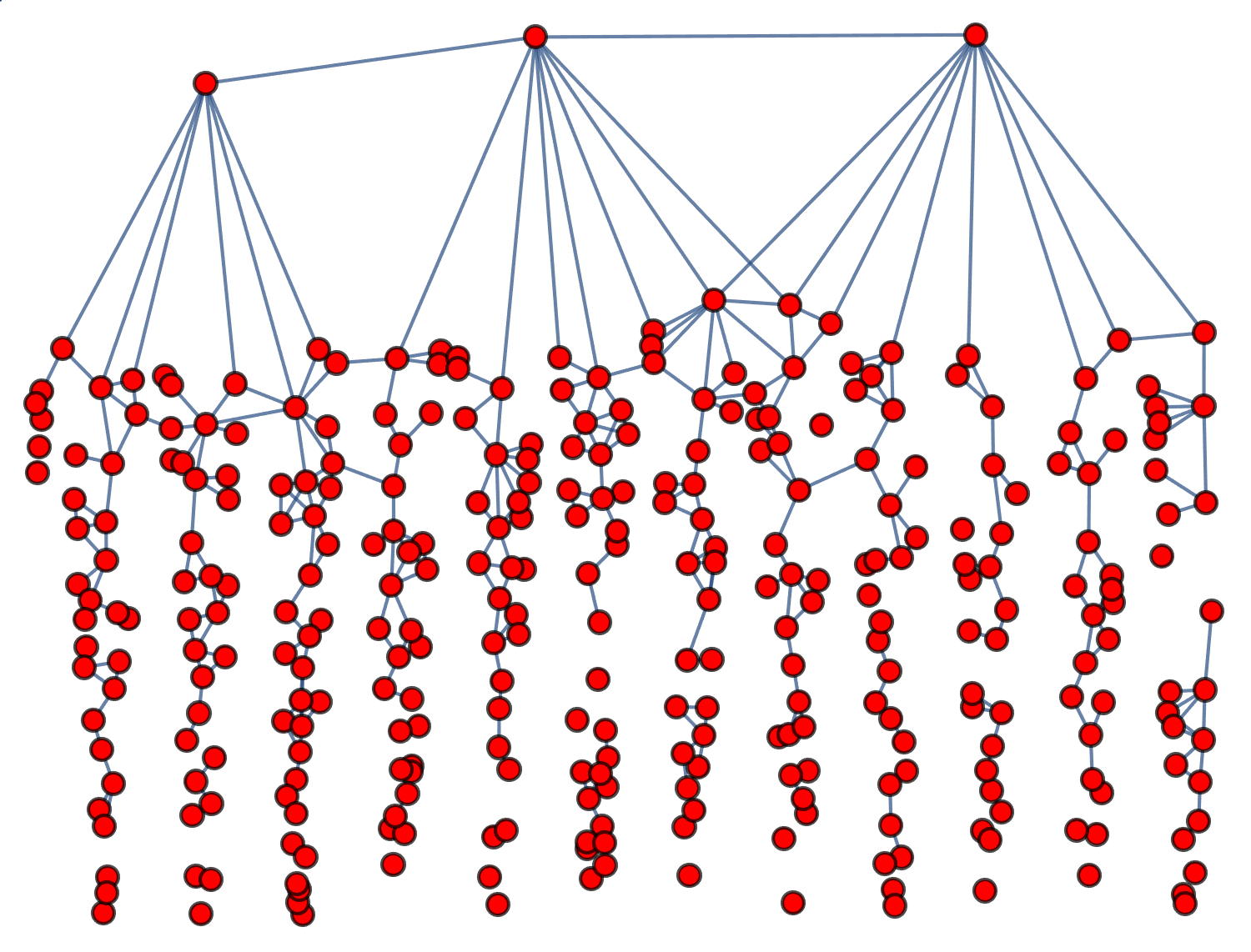}
\caption{(Top) An illustration of a data set and two manifolds with a highly nonlinear boundary. (Bottom) The simplicial complex generated for Class 1.} 
\label{fig:SineWave}
\end{figure}

Consider the data set $(X_1,X_2)$ with $X_i \subset \mathbb{R}^2$ ($i=1,2$) with the classes given by,
\begin{equation*}
    C(\mathbf{x}) = 1 \iff \sin(2\pi k x_1) \geq x_2.
\end{equation*}
For larger $k$, this data set has the property that the class boundary becomes highly nonlinear. This is illustrated in \cref{fig:SineWave} (top).
We also illustrate the constructed simplicial complex for Class 1 in \cref{fig:SineWave} (bottom) using the approach described in the previous sections. We used the standard Euclidean metric in the algorithm. Notice the simplicial structure properly reflects the nature of the boundary. 

\begin{figure}[htbp]
\centering
\includegraphics[width=0.8\columnwidth]{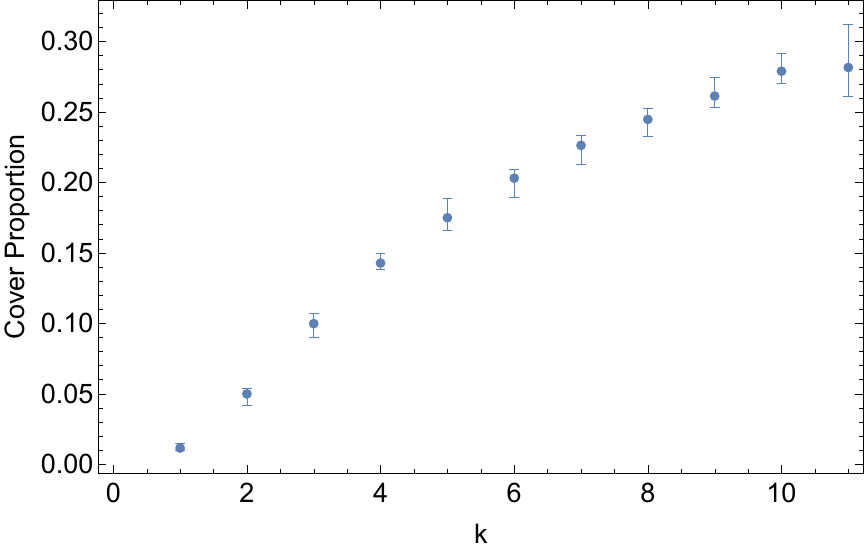}
\caption{The size of the cover increases monotonically as the complexity of the boundary between the two classes increases.}
\label{fig:IncreaseN}
\end{figure}

As $k$ increases, the boundary becomes more complex, and so also the proportion of data points in Class 1 (or Class 0) that must be used in the cover increases. This is illustrated in \cref{fig:IncreaseN}, where we also see a natural asymptote seems to occur, consistent with the limiting behavior of the geometry. 
This suggests that the size of the cover(s) of the classes, (with respect to the size classes), can be used as a natural metric on the complexity of the boundary and thus the difficulty of the learning problem. 

We set $k = 3$ and repeated the following experiment twenty times. We generated a random sample of 5000 training points and 5000 test points. We trained a deep neural network with structure $(200,200,2)$ as well as a random forest and built the topological covering. The mean accuracy with maximum and minimum over all replications are shown in \cref{tab:SineWaveAccuracy}.
\begin{table}[htbp]
\begin{tabular}{|l|c|c|}
\hline
\textbf{Method} & \textbf{Mean Acc.} & \textbf{Min/Max Acc.}\\
\hline
Topological Classifier & $0.75662$ & $(0.7468, 0.7648)$\\
\hline
Neural Network & $0.75464$ & $(0.7292, 0.778)$\\
\hline
Random Forest & $0.75883$ & $(0.7508, 0.7742)$\\
\hline
\end{tabular}
\caption{Mean accuracy and 100\% order statistics confidence intervals for the sine wave boundary classification test with $k = 3$}.
\label{tab:SineWaveAccuracy}
\end{table}
The data suggest that these three methods are largely comparable. We suspect the relatively low accuracy from all three methods is a function of the data density near the nonlinear boundary. It is well known that data density can affect the ability of algorithms in topological data analysis to recover the topological characteristics of manifolds \cite{NSW08}. We can already see this in \cref{fig:SineWave} (bottom), where the 1-skeleton of the simplicial has become disconnected. Whether and how this is affecting the random forest or neural network learning may be an area of future work. The impact of data topology on neural network learning is discussed throughout the remainder of this paper.

We repeated the experiment with $k = 8$ to see what the effect on the learning process was. The results are shown in \cref{tab:SineWaveAccuracy2}.
\begin{table}[htbp]
\begin{tabular}{|l|c|c|}
\hline
\textbf{Method} & \textbf{Mean Acc.} & \textbf{Min/Max Acc.}\\
\hline
Topological Classifier & $0.73852$ & $(0.727, 0.7514)$\\
\hline
Neural Network & $0.72631$ & $(0.7012, 0.7494)$\\
\hline
Random Forest & $0.74688$ & $(0.7354, 0.7588)$\\
\hline
\end{tabular}
\caption{Mean accuracy and 100\% order statistics confidence intervals for the sine wave boundary classification test with $k = 8$.}
\label{tab:SineWaveAccuracy2}
\end{table}
Again, we see the methods are largely comparable to each other in terms of accuracy, but the results suggest that as the boundary becomes more complex (as measured by the proportion of the data points that must be used as a cover), the ability of any method to learn the separator may decrease. This will be explored further with additional data sets. 

\subsection{Waveform Generator}
We used the waveform generator (version 1) test set available from the UCI Machine Learning Repository \cite{BS88}. The data consists of a 40 dimensional feature vector with one of three class values $0$, $1$, or $2$. The training set size consisted of 5000 samples and separate C/C++ source code is available to generate additional test samples. We used the source code to generate 1000 test samples.

\begin{figure}[htbp]
\centering
\includegraphics[width=0.95\columnwidth]{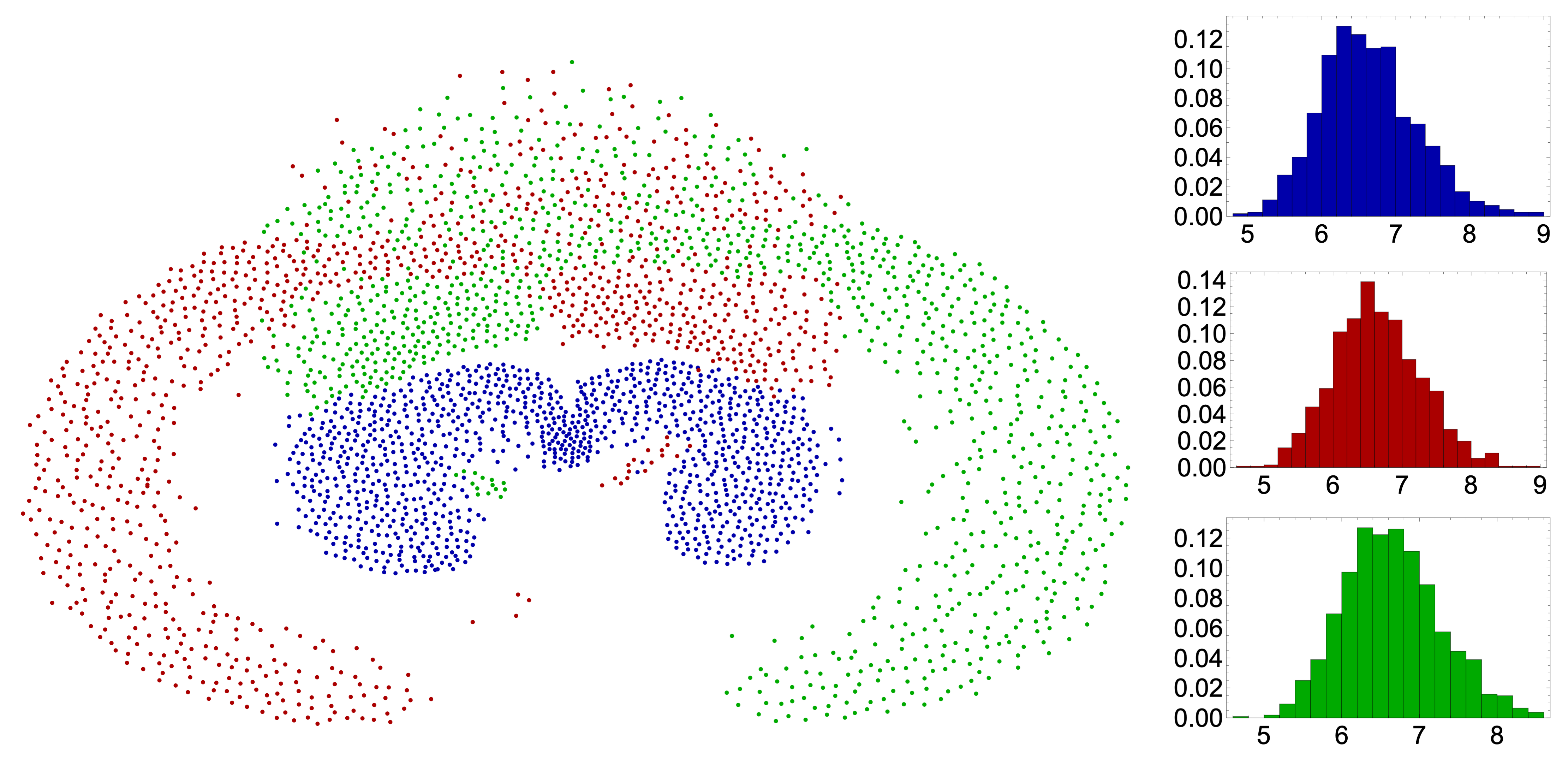}\\
\includegraphics[width=0.8\columnwidth]{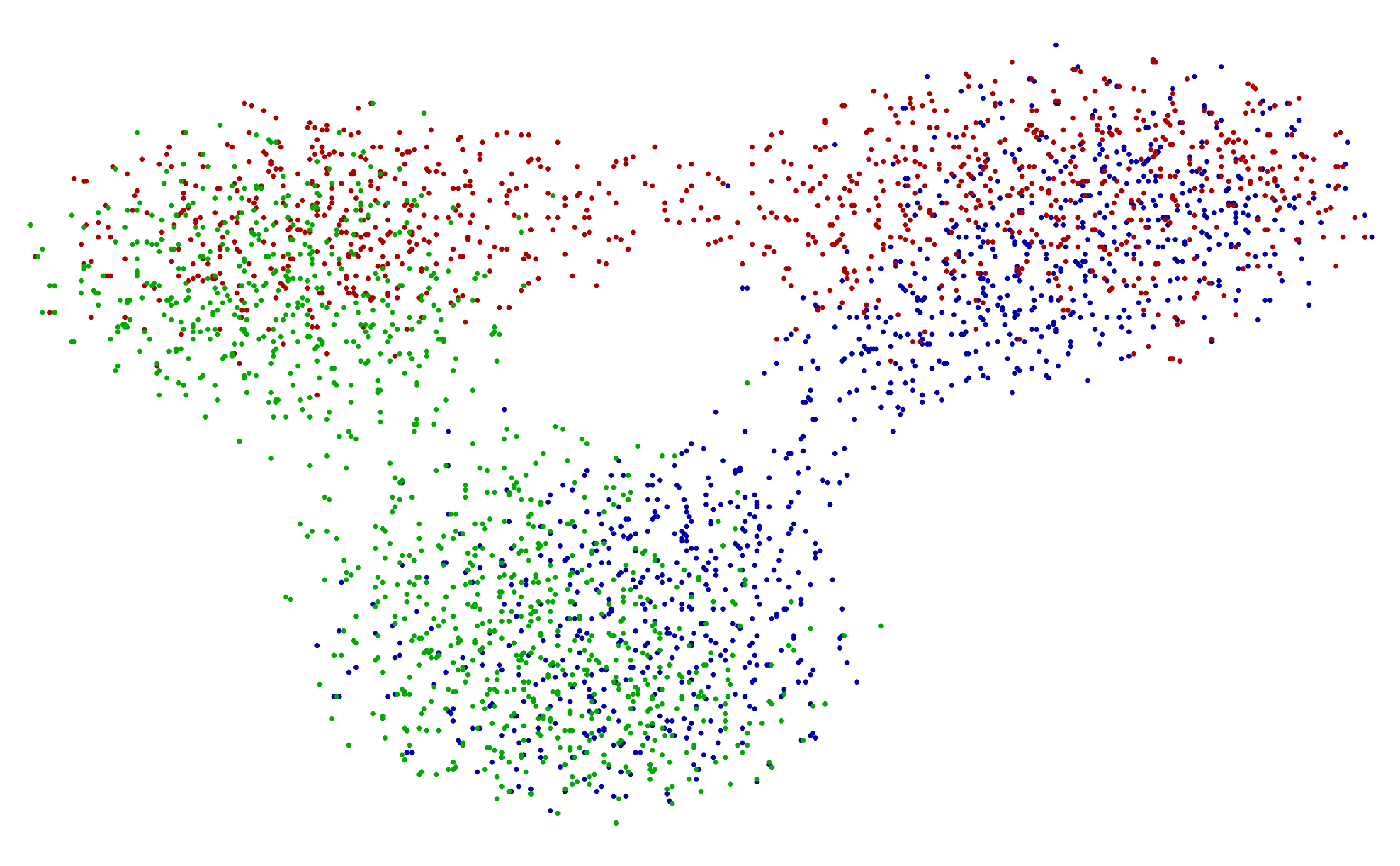}
\caption{(Top) The visualization of the joint simplicial complex of all topological covers and the histograms of the radii of the covering sets. (Bottom) Visualization of the TSNE dimensional reduction. Blue is class 0, red is class 1, green is class 2.}
\label{fig:JointWaveformComplex}
\end{figure}

We built topological coverings and simplicial complexes for the three classes of data using the standard Euclidean metric. The topology suggests the boundary between the classes is complex. As in \cref{fig:IncreaseN}, we use the proportion of the classes used to make the covering as a proxy for topological complexity. We refer to this as \textit{covering proportion}. The covering proportions of the classes are given in \cref{tab:WaveformCoverProportions}.
\begin{table}[htbp]
\centering
\begin{tabular}{|l|c|c||c|c|}
\hline
\textbf{Class} & \textbf{Size} & \textbf{Cover Size} & \textbf{Cover Prop.} & \textbf{Connected}\\
\hline
Class 0 & 1692 & 1072 & 0.63 & True\\
\hline
Class 1 & 1653 & 1016 & 0.61 & True\\
\hline
Class 2 & 1655 & 1079 & 0.652 & True\\
\hline
\end{tabular}
\caption{The cover sizes of the three classes in the waveform generator data.}
\label{tab:WaveformCoverProportions}
\end{table}
We can determine that all three simplicial complexes are connected (i.e., $\beta_0 = 1$ for all simplexes). However, because the covers are comparatively large, it is difficult to generate a general Betti sequence for the simplicial complexes. The fact that so much of the data are used to build the minimal covering suggests a complex boundary structure. Even without the explicit Betti sequences, we can explore the nature of the boundary by generating a joint simplicial complex for all classes using the covers and visualizing the result using a graph visualization algorithm. We define a graph (skeleton) $\overline{G} = (\overline{V},\overline{E})$  that combines all covers using the edge relation,
\begin{equation*}
    \{i_j, k_l\} \in \overline{E} \iff B_{r_{i_j}}(x_{i_j}) \cap B_{r_{k_l}}(x_{k_l}),
\end{equation*}
where $i$ and $k$ are indexed over class and $j$ and $l$ are indexed over the cover elements of the respective classes. This graph has as sub-graphs the graphs $G_i = (C_i^*,E_i)$ but also includes edges between the covers. We show the joint simplicial complex in \cref{fig:JointWaveformComplex} (top) using a  spring-electrical layout in which vertices are treated as charged objects connected by edges treated as springs \cite{HS15}. This layout option is provided natively in Mathematica. Note that edges in the skeleton are removed from the visualization for clarity.
Each point in \cref{fig:JointWaveformComplex} is the center of a ball in the topological covers, and as such a witness point. Since these points are designed to cover their respective manifolds, we see that not all of them are necessarily close to a class boundary. This is illustrated by the histograms of the radii of the covering elements in the top right. By way of comparison, we show the TSNE \cite{HR02} projection of the points in the cover. The projection of the simplex seems to provide substantially more information, showing that the topological spaces form a kind of nested structure with the possibility that there are high-dimensional voids where the topological spaces pass through each other. 

\begin{figure}[htbp]
\centering
\includegraphics[width=0.45\columnwidth]{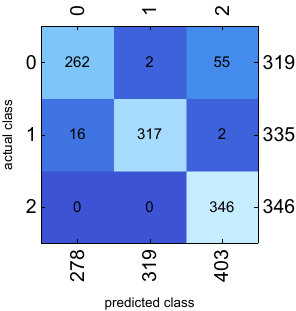}
\includegraphics[width=0.45\columnwidth]{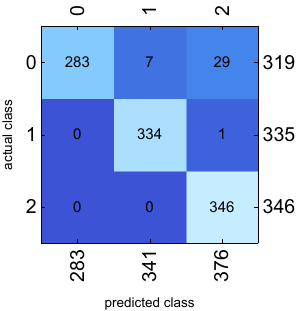}\\
\includegraphics[width=0.45\columnwidth]{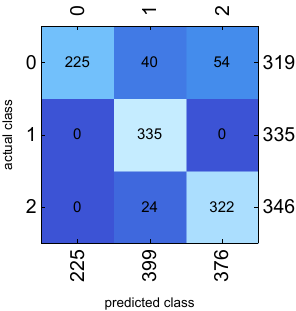}
\includegraphics[width=0.45\columnwidth]{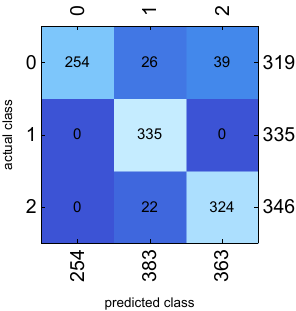}
\caption{Confusion matrices for the waveform data set. (Top Left) Topological classifier. (Top Right) Random forest classifier. (Bottom Left) Deep neural network classifier. (Bottom Right) Shallow neural network classifier.}
\label{fig:WaveFormConfusionMatrix}
\end{figure}

We tested the topological classification algorithm against a deep neural network classifier with structure $(500,100,3)$, a shallow neural network classifier with structure $(3000,3)$ and a random forest, using the 1000 test samples we generated. The accuracy results are shown in \cref{tab:WaveformAccuracy}.
\begin{table}[htbp]
\begin{tabular}{|l|c|}
\hline
\textbf{Classifier} & \textbf{Accuracy}\\
\hline
Topological & $92.5 \pm 0.8\%$\\
\hline
Deep Neural Network & $88.2 \pm 1.0\%$\\
\hline
Shallow Neural Network & $91.3 \pm 0.9\%$\\
\hline
Random Forest & $96.3 \pm 0.6 \%$\\
\hline
\end{tabular}
\caption{Accuracy table for waveform classification data set. Uncertainties are computed automatically by Mathematica and correspond to one standard deviation.}
\label{tab:WaveformAccuracy}
\end{table}
Confusion matrices for the four classifiers are shown in \cref{fig:WaveFormConfusionMatrix}.
Intriguingly, the random forest classifier outperforms the topological classifier, which in turn outperforms the deep neural network classifiers, but is statistically identical to the shallow neural network classifier. It is possible that increasing the width of the shallow neural network would improve the score of the neural network. We hypothesize that the complexity of the boundary between the classes, as illustrated by the topological analysis, is causing a challenge in the learning process of the deep neural network. We explore this further with additional data sets.

\subsection{MNIST}
We built a topological model of the MNIST data set \cite{D12} to illustrate additional features of the topological approach. We used the \texttt{ImageDistance} metric in Mathematica as the metric. We compare the classification results from the topological classifier, a random forest model and LeNet \cite{LBYH98}, an early convolutional neural network (CNN) trained specifically on MNIST. 

\begin{figure}[htbp]
\centering
\includegraphics[width=1\columnwidth]{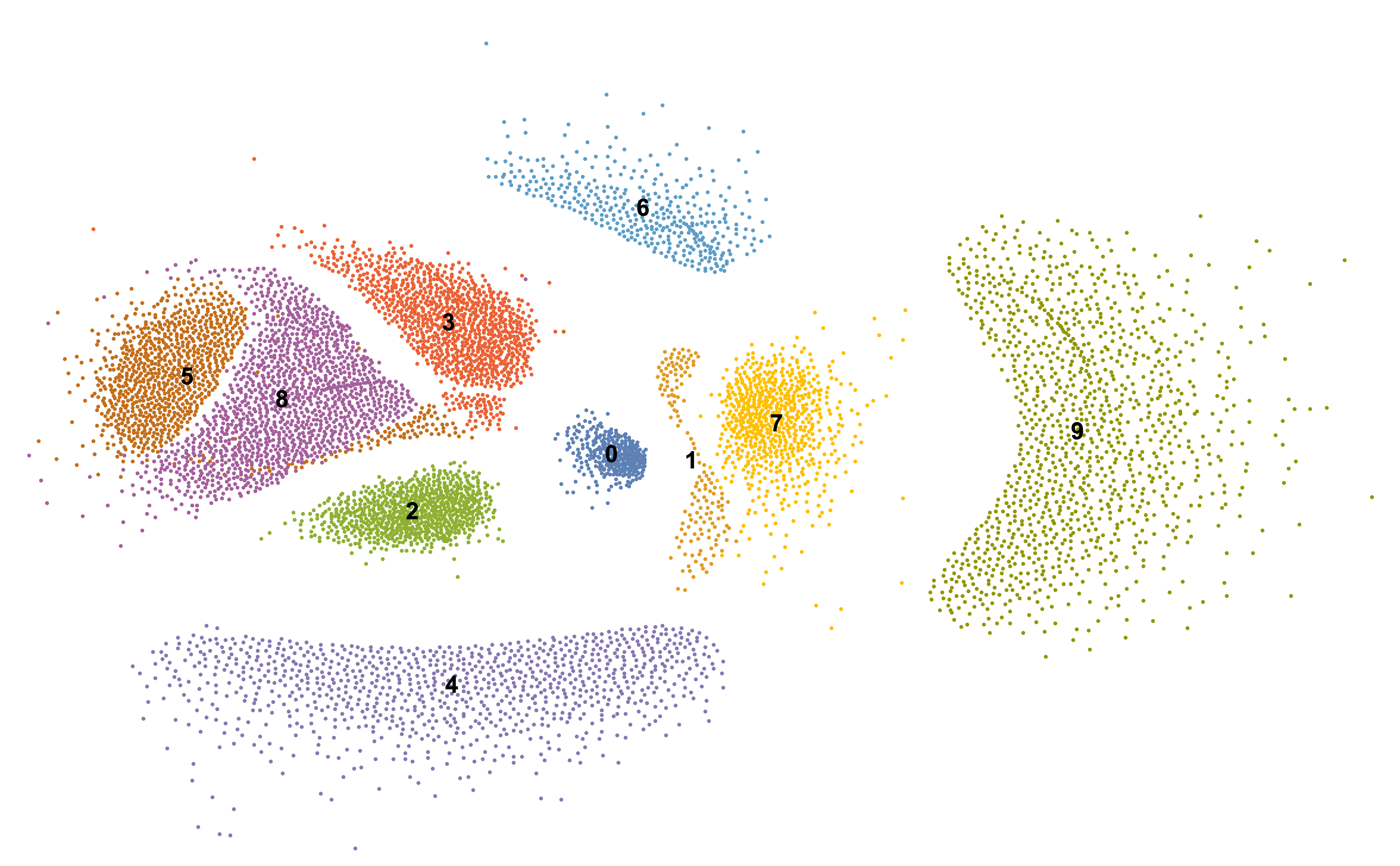}
\caption{A visualization of the MNIST data using the simplicial complex models of the manifolds.}
\label{fig:MNISTManifolds}
\end{figure}

Basic topological information on MNIST is shown in Table \ref{tab:MNISTTop}.
\begin{table}[htbp]
\begin{tabular}{|l|c|c|}
\hline
\textbf{Digit} & \textbf{Connected} & \textbf{Cover Prop.}\\
\hline
0 & {True} & 0.053 \\
\hline
 1 & {True} & 0.023 \\
 \hline
 2 & {True} & 0.137 \\
 \hline
 3 & {True} & 0.172 \\
 \hline
 4 & {True} & 0.17 \\
 \hline
 5 & {True} & 0.176 \\
 \hline
 6 & {True} & 0.067 \\
 \hline
 7 & {True} & 0.11 \\
 \hline
 8 & {True} & 0.232 \\
 \hline
 9 & {True} & 0.195\\
 \hline
\end{tabular}
\caption{The basic topological information from MNIST suggests that the manifolds on which digits 0 and 1 reside are the easiest to separate, while digit 8 may be the hardest.}
\label{tab:MNISTTop}
\end{table}
The computed simplicial complexes for the classes of data are large, with skeletons containing between 157 and 1358 vertices and between 10,334 and 893,250 edges. This makes it impossible to compute exact topological information for all classes, beyond the fact that all inferred manifolds are path connected, as shown in \cref{tab:MNISTTop}. The raw data could be analyzed using standard techniques from topological data analysis, e.g., persistent homology with a secondary witness complex \cite{DC04}, but this would eliminate the possibility of extracting class-level features relevant to the decision boundaries. While developing techniques for handling the potentially large inferred simplicial complexes resulting from our approach is left to future work, we can use the simplex skeletons to generate a visual representation of the manifolds. This is shown in \cref{fig:MNISTManifolds}.
This visualization was generated using a spring-electrical layout (from Mathematica) as described in the previous section. However, to increase the clarity of the visualization, we randomly removed half of the edges linking different classes in the skeleton (graph) of the joint simplicial complex prior to visualization. The topological covering proportions are smallest for digits 0 and 1, suggesting they are the simplest digits, topologically speaking, while digits 8 and 9 require the most information to generate the cover, suggesting these digits may be more difficult to separate, most likely due to the similarity between 5 and 8 and possibly 7 and 9 (see \cref{fig:MNISTManifolds}). 

\begin{figure}
\centering
\includegraphics[width=0.9\columnwidth]{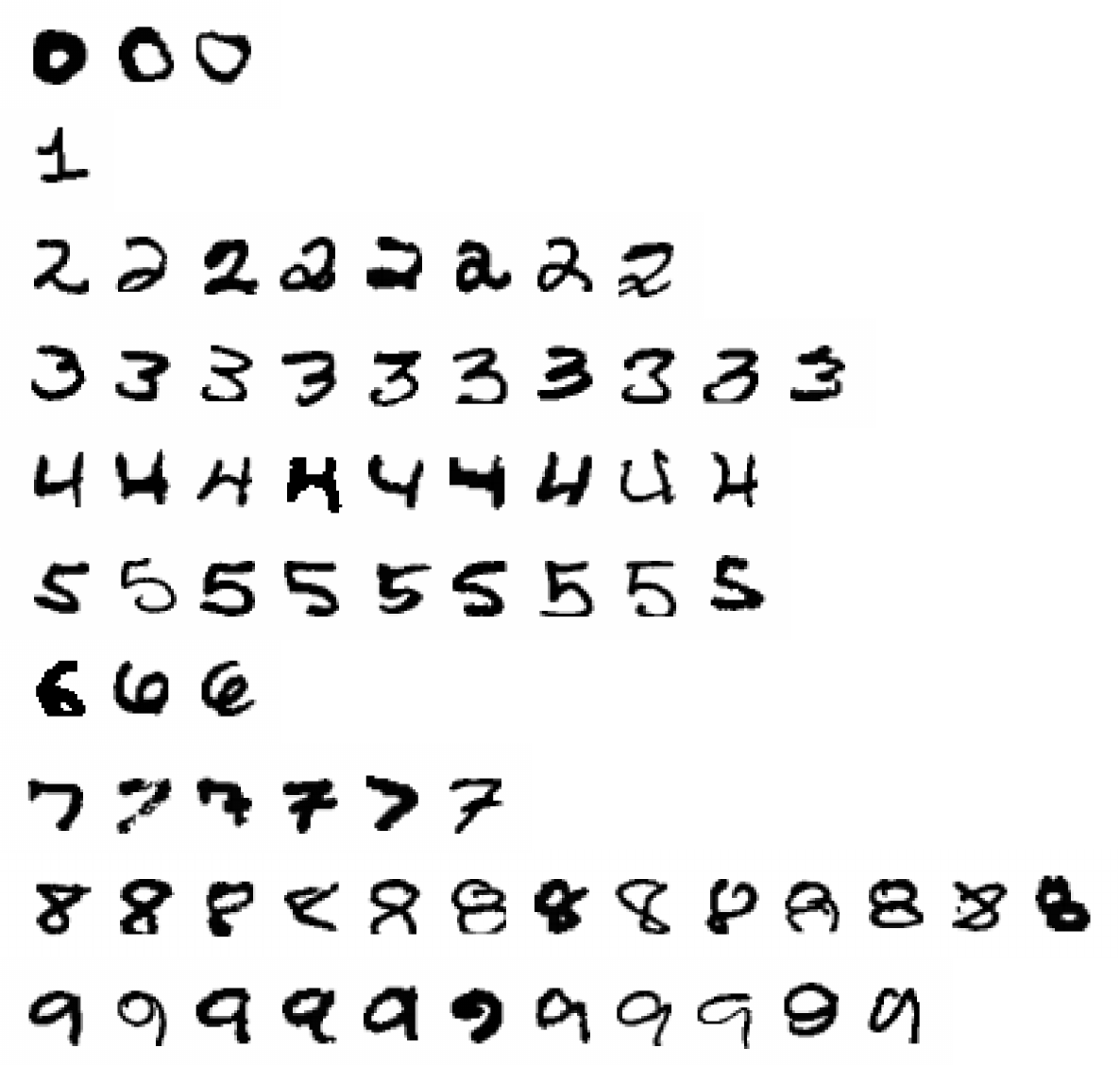}\\    
\caption{The images with the largest radii are those that are farthest from the boundary between classes and represent archetypal class elements.}
\label{fig:Normal}
\end{figure}

The radii of the balls in the covering can be used to generate additional information. By selecting the covering images with the smallest radii (in the $1^\text{st}$ percentile), we can identify those images that are close to the boundary, and thus causing confusion. 
\begin{figure}
\centering
\includegraphics[width=0.9\columnwidth]{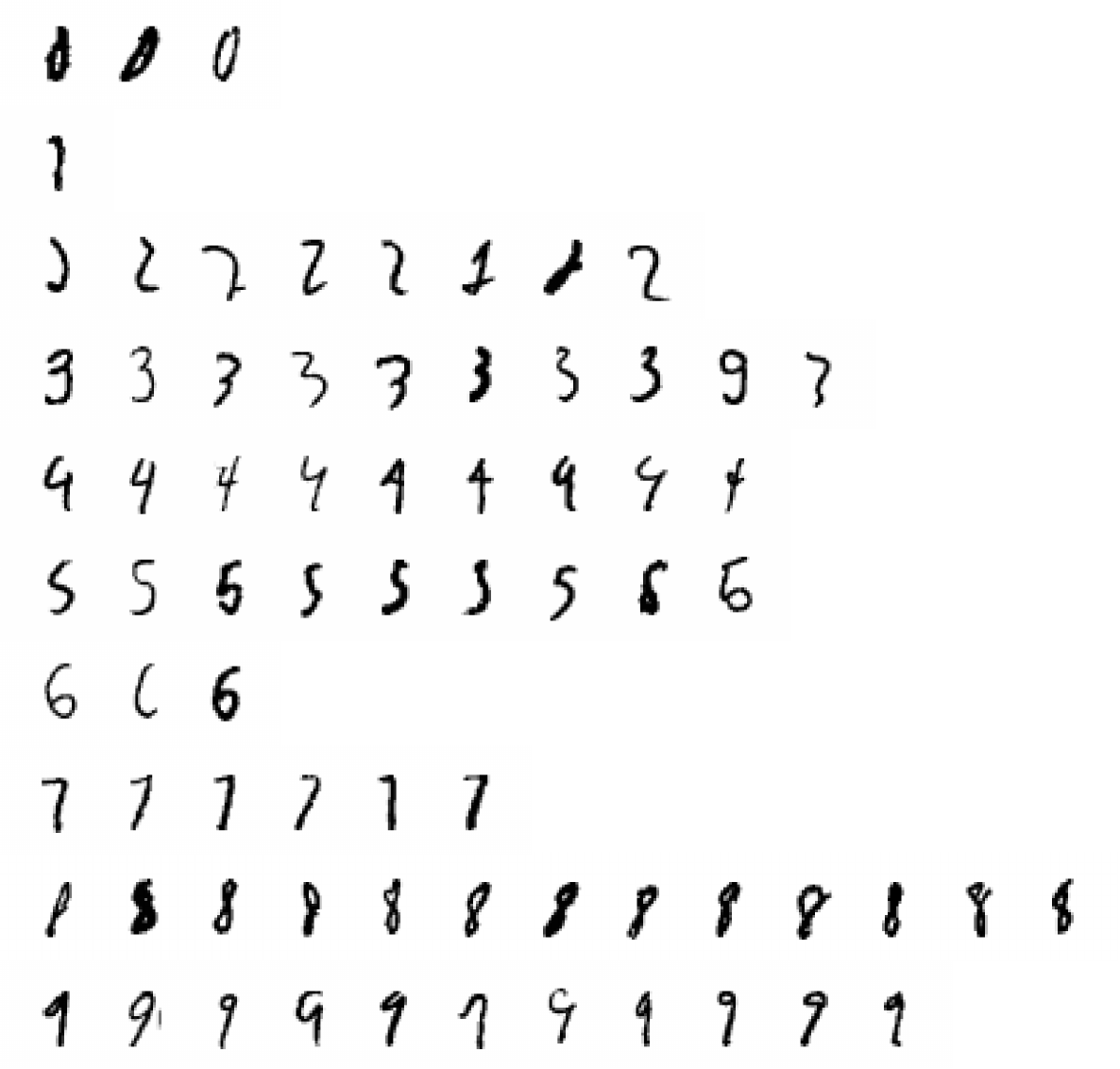}\\    
\caption{The images with the smallest radii are those that are closest to the boundary between classes and represent archetypal points of confusion.}
\label{fig:Weird}
\end{figure}
These are shown in \cref{fig:Weird}. Likewise, cover elements with large radii (in the $99^\text{th}$ percentile) represent points far from decision boundaries and are archetypal class elements. This is shown in \cref{fig:Normal}. 

A classifier accuracy comparison is shown in \cref{tab:MNISTAccuracy}, using the standard MNIST test set of 10,000 samples.
\begin{table}[htbp]
\begin{tabular}{|l|c|}
\hline
\textbf{Classifier} & \textbf{Accuracy}\\
\hline
Topological & $95.27 \pm 0.21\%$\\
\hline
LeNet Network & $98.48 \pm 0.12\%$\\
\hline
Random Forest & $95.6 \pm 0.21 \%$\\
\hline
\end{tabular}
\caption{Accuracy table for MNIST data set. Uncertainties are computed automatically by Mathematica and correspond to one standard deviation.}
\label{tab:MNISTAccuracy}
\end{table}
As we can see, LeNet outperforms both the topological approach and the random forest, which are both statistically indistinguishable. We hypothesize that this is because the CNN is better able to approximate the manifolds of the data sets, and thus the nonlinear boundaries. 
\begin{figure}[htbp]
\centering
\includegraphics[width=0.65\columnwidth]{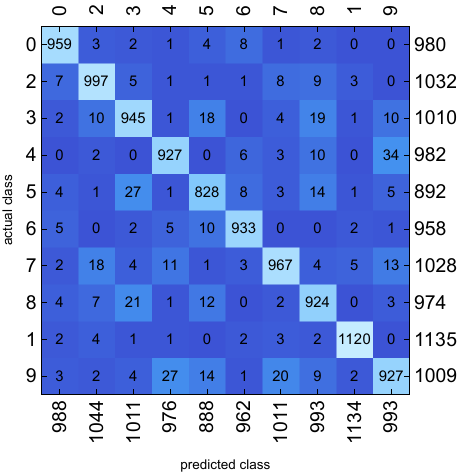}\\
\includegraphics[width=0.65\columnwidth]{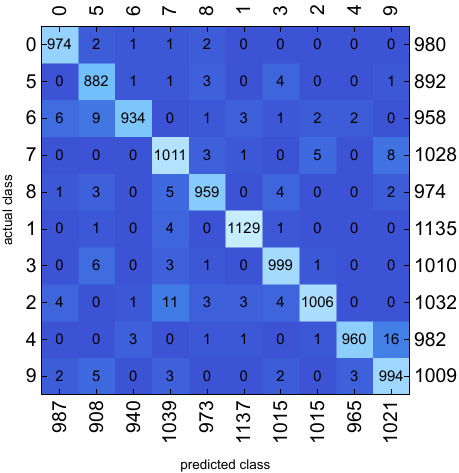}\\
\includegraphics[width=0.65\columnwidth]{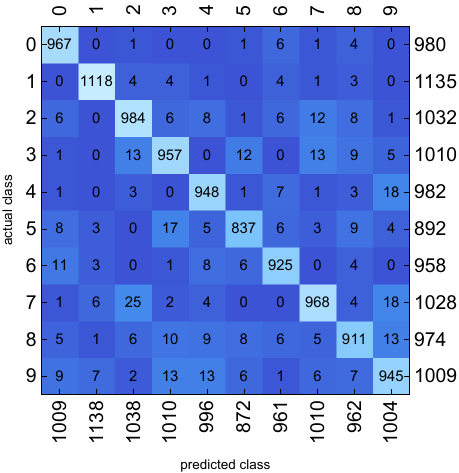}
\caption{Confusion matrices for the MNIST data set. (Top) Topological classifier. (Middle) Neural network classifier. (Bottom) Random forest classifier.}
\label{fig:MNISTConfusionMatrix}
\end{figure}
The confusion matrices for these experiments are shown in \cref{fig:MNISTConfusionMatrix}. Investigation of the relationship between topological structure and CNN's is reserved for future work.

\subsection{HEPMASS}
To test the scalability of the cover generation and topological classification approach, we used the open source HEPMASS dataset \cite{BCFS16}, which is composed of 10.5M Monte Carlo simulations of high-energy particle collisions. The data are divided into 7M simulations where the signal particle has mass equal to a nominal 1000 with the remaining examples having variable mass. We used the 7M samples with fixed mass 1000. The data resides in a 27 dimensional Euclidean space. The two classes correspond to a relevant particle being present or absent. Unlike the previous sections, topological coverings were generated using a C++ implementation. (See data availability statement for source code access.)

Independent testing suggests the data are amenable to classification by DNN \cite{M21}. To explain this, we constructed the two simplicial complexes $H_0$ and $H_1$ using open balls for the topological covering. Because the data set is large (by the standards of most topological data analysis algorithms), we cannot compute a complete set of topological features. Graph-theoretic analysis suggests that the two topological spaces $\mathcal{T}_1$ and $\mathcal{T}_0$ are largely connected. The space $\mathcal{T}_1$ consists of a giant component with a second small disconnected component, which maps to an isolated vertex in the simplicial complex $H_1$. The topological space $\mathcal{T}_0$ is path connected. Information on the topological structure is shown in \cref{tab:HEPMASSGross}.   
\begin{table}[htbp]
\centering
\begin{tabular}{|c|c|c|c|}
\hline
\textbf{Class} & \textbf{Sample Size} & \textbf{Cover Prop.}  & \textbf{$\beta_0$}\\
\hline
Class 1 & 3,500,879 & 0.84  & 2 (1 dimension 0)\\
\hline
Class 0 & 3,499,129 & 0.58 & 1\\
\hline
\end{tabular}
\caption{High level topological features of the HEPMASS data set. Here, 0 dimensional simplexes are isolated vertices in $H_1$ or $H_0$.}
\label{tab:HEPMASSGross}
\end{table}  
The proportion of the data used to create the covering is  large compared, with 84\% of the Class 1 data used as witness points to form the cover. This suggests a boundary with complex nonlinear structure. Unfortunately, the data set is too large to compute exact Betti numbers, but we can visualize the data by using iGraph's implementation of large graph layout \cite{CN+06}. 

\begin{figure}[htbp]
\centering
\includegraphics[width=0.65\columnwidth]{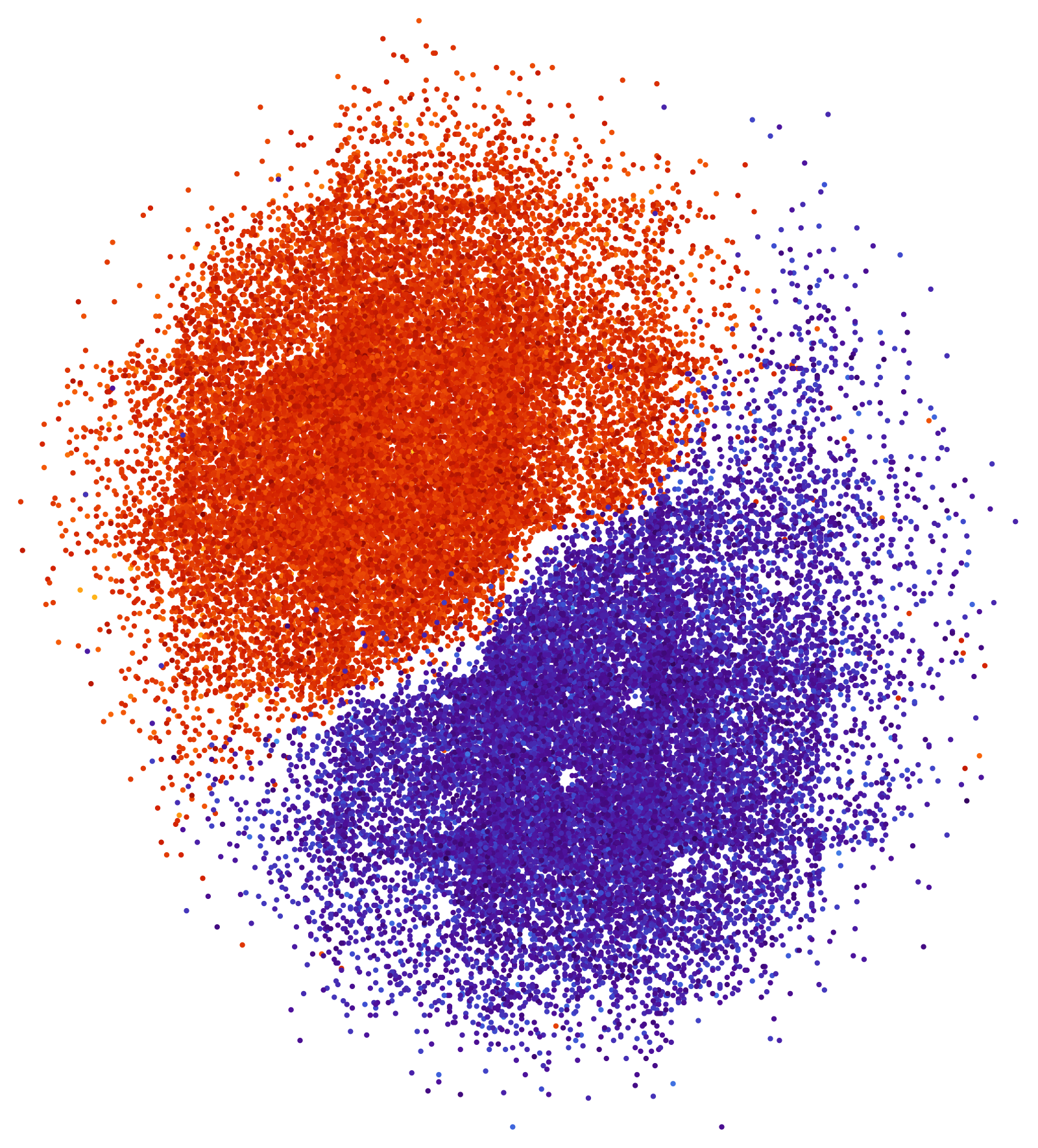}
\caption{A visualization of a witness simplicial complex generated from the HEPMASS particle physics data set \cite{BCFS16} shows that this data should be easily separable by a simple nonlinear manifold. A DNN will project the high-dimensional data into a lower dimensional representation with a simple separation, much like this energy-minimizing visualization does.}
\label{fig:HEPMASSCover}
\end{figure}

To speed up computation, 150,000 balls from the topological coverings defining $S_0$ and $S_1$ were chosen at random. This effectively produced a smaller witness complex  \cite{DC04} containing 300,000 points. We refer to this as the witness cover. The resulting witness complex skeletons are visualized  in \cref{fig:HEPMASSCover}.
The visualization suggests that the data are nonlinearly separable, with a complex separating manifold between them. In some sense, this is a high-dimensional analogue of the boundary illustrated in \cref{sec:SineWave}.

We used 20 random samples of the HEPMASS testing data set, each of size 20,000 and applied the topological classification algorithm. Results of this experiment are shown in  \cref{tab:Results}.
\begin{table}[htbp]
\centering
\begin{tabular}{|l||c|c|}
\hline
\textbf{Measures} & \textbf{Mean} & \textbf{(Min, Max)} \\
\hline
\hline
 {Total Accuracy} & $0.881$ &{$(0.879, 0.885)$}\\
 \hline
 {True Positive} & $0.897$ & {$(0.893,0.901)$} \\
 \hline
 {False Positive} & $0.134$ & {$(0.13,0.139)$} \\
 \hline
 {F1 Score} & $0.883$ & {$(0.861,0.887)$} \\
 \hline
 {Out of Cover Prop.} & $0.547$ & {$(0.54,0.553)$} \\
 \hline
 {Confused Cover Prop.} & $0.00326$ & {$(0.00275,0.00383)$} \\
 \hline
\end{tabular}
\caption{Results show mean total accuracy of 88\% with similar true positive and F1 scores. Interestingly, 55\% of the test data (on average) was outside the cover, meaning substantial generalization was required. Only 0.3\% of the test data fell close enough to the boundary to lie in both covers.}
\label{tab:Results}
\end{table}
The data suggest that the cover accurately respects the topological structure of $\mathcal{T}_1$ and $\mathcal{T}_0$ including the boundary, with only $0.33\%$ of test samples confused between the two covers. However, testing samples frequently fall outside the specific boundaries of the cover, which would require generalization from a neural network classifier. This may be a result of our use of a witness cover. Based on these results, we expect to see high-quality separation from a DNN. To test this hypothesis, we used a DNN with architecture $(128,64,32,2)$ ending in a softmax classifier and tested on 20 replications of 20,000 test points chosen at random. Results are shown in \cref{tab:Results2} and are consistent with \cite{M21}.
\begin{table}[htbp]
\centering
\begin{tabular}{|l||c|c|}
\hline
\textbf{Measures} & \textbf{Mean} & \textbf{(Min, Max)} \\
\hline
\hline
 {Total Accuracy} & $0.89$ &{$(0.888, 0.8886)$}\\
 \hline
 {True Positive} & $0.89$ & {$(0.878, 0.891)$} \\
 \hline
 {False Positive} & $0.12$ & {$(0.116, 0.129)$} \\
 \hline
 {F1 Score} & $0.88$ & {$(0.88,0.887)$} \\
 \hline
\end{tabular}
\caption{Results show mean total accuracy of 88\% with similar true positive and F1 scores. The false positive rate is slightly lower than the false positive rate of the topology-based classifier.}
\label{tab:Results2}
\end{table}
The neural network outperforms the topology-based classification method, but only by 1\% on average. We attribute this to (i) improved ability to model the separating boundary and (ii) better generalization. However, the topological simplicity of the underlying data clearly explains a substantial portion of the DNN's success. 

\subsection{Undersea Acoustic Data Set}
\begin{figure}[htbp]
\centering
\includegraphics[width=0.85\columnwidth]{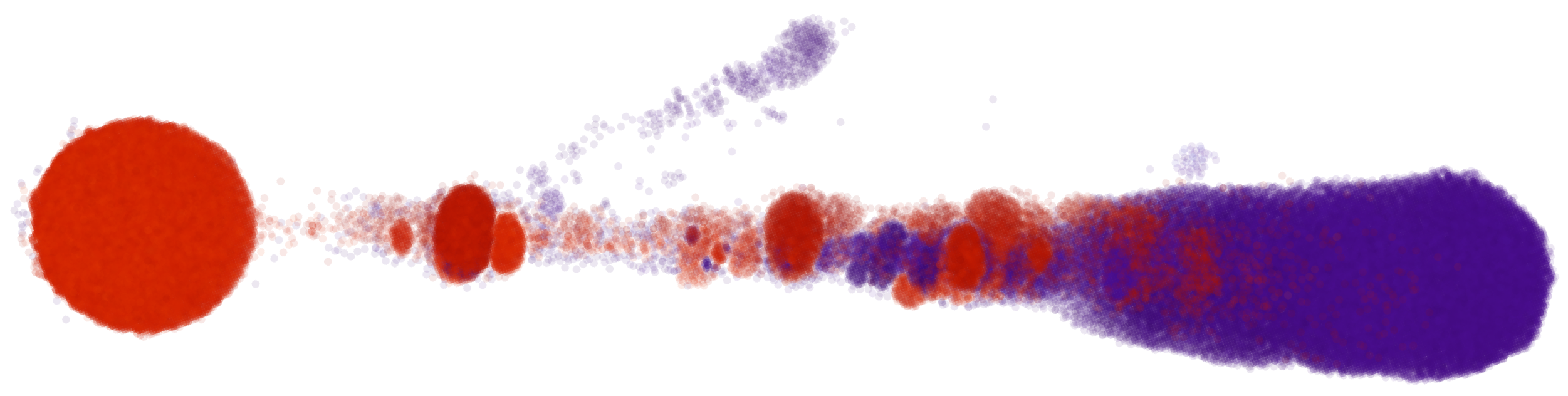}
\caption{Visualization of the witness simplicial complex of the two click classes of \textit{Lagenorhynchus obliquidens}.}
\label{fig:DolphinEmbedding}
\end{figure}

We performed a similar analysis on open-source data provided by the Scripps Institute \cite{F21} consisting of Fourier transforms of undersea acoustic data. There were $\approx 1.5M$ samples from \textit{Lagenorhynchus obliquidens}. Class 1 was composed of Type A clicks and Class 0 was composed of Type B clicks. This data set is fully described and analyzed using a DNN in \cite{F21}, where it is shown empirically the data are amenable to classification by DNN. Topological analysis indicates that the classes lie on path connected manifolds residing in 181 dimensional Euclidean space. There are no disconnected elements, though again the derived simplicial complexes are too large (edge dense) to allow for complete topological investigation. A summary of the gross topological properties of $S_0$ and $S_1$ is given in \cref{tab:ScrippsGross}. 
\begin{table}[htbp]
\centering
\begin{tabular}{|c|c|c|c|}
\hline
\textbf{Class} & \textbf{Sample Size} & \textbf{Cover Prop.}  & \textbf{$\beta_0$}\\
\hline
Class 1 & 798,116 & 0.40  & 1\\
\hline
Class 0 & 679,894 & 0.55 & 1\\
\hline
\end{tabular}
\caption{High level topological features of the Scripps data set. Both $S_1$ and $S_0$ have one component ($\beta_0 = 1)$ implying the underling manifolds on which Type A and Type B clicks reside are path connected.}
\label{tab:ScrippsGross}
\end{table} 
As before, we created a visualization of the joint simplex, shown in \cref{fig:DolphinEmbedding}.  
Interestingly, this visualization suggests the data are more complex than the corresponding HEPMASS data. Further topological analysis is required to prove this is true. 

To test the goodness of a covering generated by a subset of this data, we removed 20,000 samples from each class and rebuilt a topological covering with the remaining data. The data suggest that 96\% accuracy can be achieved using the topological classifier, with only 4.5\% of test samples confused between the two classes. In this test, 19\% of the test data was outside the cover. Complete results are shown in \cref{tab:Results3}.
\begin{table}[htbp]
\centering
\begin{tabular}{|l||c|c|}
\hline
\textbf{Measures} & \textbf{Value}\\
\hline
\hline
 {Total Accuracy} & $0.96$ \\
 \hline
 {True Positive} & $0.96$  \\
 \hline
 {False Positive} & $0.042$ \\
 \hline
 {F1 Score} & $0.96$  \\
 \hline
 {Out of Cover Prop.} & $0.19$  \\
 \hline
 {Confused Cover Prop.} & $0.045$  \\
 \hline
\end{tabular}
\caption{Results show a total accuracy of 96\% with 96\% true positive and F1 score 0.96. Interestingly, 19\% of the test data (on average) were outside the cover, meaning some generalization was required. Only 4.5\% of the test data fell close enough to the boundary to lie in both covers.}
\label{tab:Results3}
\end{table}
A DNN with architecture $(1024, 128, 64, 32, 2)$ ending in a softmax classifier was also trained on the data. The total accuracy using this DNN was also 96\%, with a slightly lower false positive rate than the topological approach, but not considerably so. This suggests the DNN may be slightly better at generalizing the complex boundary structure than the open sets forming the covering of the topology. The full test results are in \cref{tab:Results4}.
\begin{table}[htbp]
\centering
\begin{tabular}{|l||c|c|}
\hline
\textbf{Measures} & \textbf{Value} \\
\hline
\hline
 {Total Accuracy} & $0.96$ \\
 \hline
 {True Positive} & $0.96$  \\
 \hline
 {False Positive} & $0.036$ \\
 \hline
 {F1 Score} & $0.96$ \\
 \hline
\end{tabular}
\caption{Results show a total accuracy of 96\% with similar true positive and F1 scores. The false positive rate is lower than the topology-based classifier, suggesting better generalization.}
\label{tab:Results4}
\end{table}
As before, we hypothesize that the topological simplicity of the underlying data explains a substantial portion of the DNN's success. We attempt to investigate this hypothesis further in the next sections.

\section{Topological Implications for Learning Neural Networks: Tile Model}\label{sec:Carpet}
\begin{figure}[htbp]
\centering
\includegraphics[width=0.9\columnwidth]{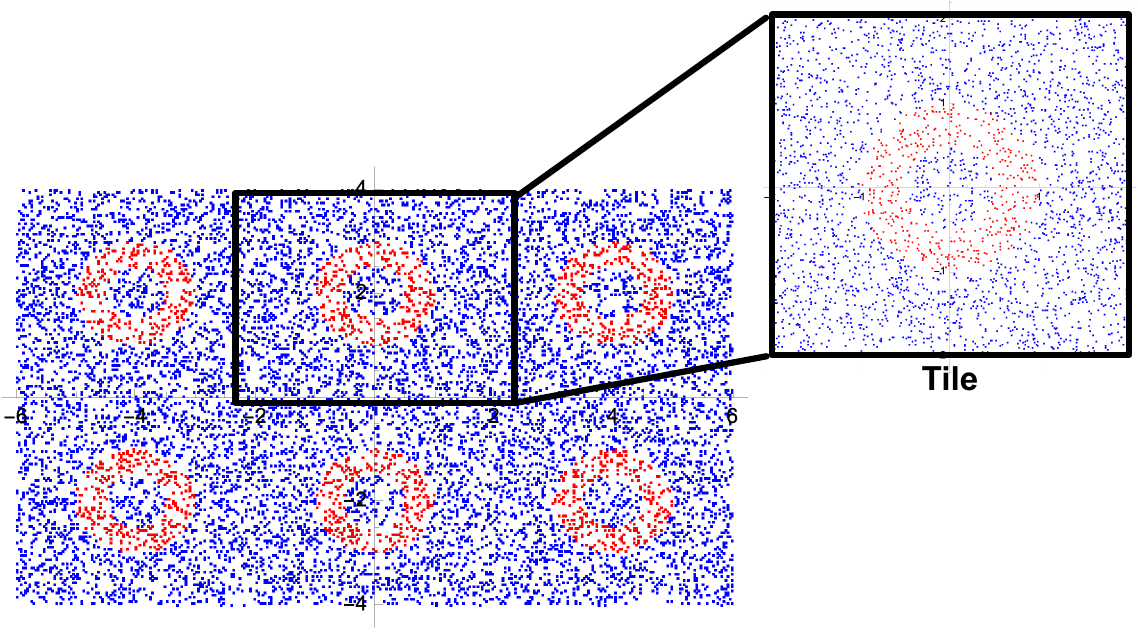}
\caption{A single tile is tessellated to construct a two-class data set in which both classes have a controllable number of holes.}
\label{fig:Tile}
\end{figure}

To test the impact of topological complexity on learning, we constructed a custom two-dimensional data set. This was accomplished by tessellating the two class data set shown in \cref{fig:Tile}. 
In \cref{fig:Tile}, Class 1 is shown in red and lies on (the union of) several annuli, while Class 0 (shown in red) is the complement of Class 1 in the plane. By way of example, an $n\times m$ tessellation would have $nm$ holes and $nm$ components in the manifold on which Class 1 lies and $nm$ holes and $2nm$ components in the manifold on which Class 0 lies. Thus, the exact Betti numbers for the data follow from the construction.

As before, we evaluated the topological classification algorithm, a random forest and feedforward neural networks. All neural networks had structure $(k, k, k, 2)$, where $k$ increased with the topological complexity of the data set. \cref{tab:NetSizes} shows the neural network sizes as a function of the number of holes in the data.
\begin{table}[htbp]
\centering
\begin{tabular}{|c|c|}
\hline
\textbf{Number of Holes} & \textbf{DNN Size ($k$)}\\
\hline
1 & 250 \\
\hline
 2 & 500 \\
 \hline
 3 & 750 \\
 \hline
 4 & 1000 \\
 6 & 1500 \\
 \hline
 8 & 2000 \\
 \hline
 9 & 2250 \\
 \hline
 12 & 3000 \\
 \hline
 16 & 4000\\
 \hline
\end{tabular}
\caption{The sizes of the layers in the neural networks used to build a classifier from the tiled data set.}
\label{tab:NetSizes}
\end{table}

\begin{figure}[htbp]
\centering
\includegraphics[width=0.95\columnwidth]{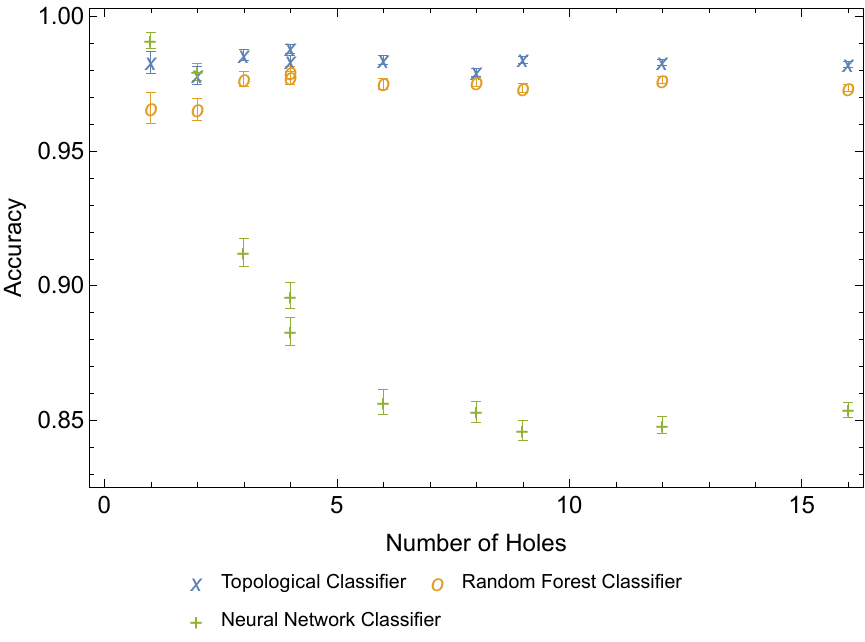}
\caption{The classification accuracy on tiled data sets as a function of the number of holes in Class 1. This shows that as topological complexity increases, learning suffers.}
\label{fig:TileResults}
\end{figure}

We built test data sets by generating 1000 random points in the base tile and tessellating these points in the same way the corresponding training data set was generated. Results for all three classifiers as a function of the number of holes in Class 1 are shown in \cref{fig:TileResults}.
In this data set, as topological complexity increases, as measured by the number of holes in Class 1 (or Class 0), we see that neural network learning suffers. It is interesting to note that the random forest approach does not suffer a similar learning failure. The topological classification algorithm works well, consistently outperforming the random forest method. This work supports our hypothesis that the topological complexity of the data has an impact on DNN learning. In the next section, we further validate this hypothesis and show that this behavior is repeatable for a data set derived from a children's game, Math Dice Jr.  

\section{Topological Implications for Learning a Children's Game}
\label{sec:MathDiceJr}
We now analyze a data set arising from a children's game that is amenable to complete topological analysis. That is, we can compute all Betti numbers and use them to evaluate the impact of complex topological structures on learning. 

The game \textit{Math Dice Jr.} is a fundamentally simple game. Three ordinary dice and two six sided dice containing only the numbers $1$, $2$ and $3$ are rolled along with a dodecahedron. The objective is to use only addition and subtraction on the values shown on the six-sided die to arrive at the value on the dodecahedral die. For arbitrary numbers of dice, this problem can be shown to be equivalent to the $\mathrm{NP}$-hard subset sum problem. In the experiments below, we consider Math Dice with 2–5 six-sided dice (or 3–6 dice overall). In all cases except the 6 dice case, we assume the 6-sided dice are ordinary (having numbers 1–6). In the 6 dice case, we model the Math Dice Jr. game. With $n$ dice, each dice roll can be represented as a point $\mathbf{x} \in \mathbb{R}^n$. For our analysis, we define $\mathbf{x}\in X_1$ if and only if all $n-1$ six-sided dice can be used to recover the value on the $n^\text{th}$ die. This divides the rolls into two classes, $(X_0,X_1)$. 

All Math Dice Jr. games with $n$ dice can be solved using a simple integer programming problem,
\begin{equation*}
\left\{
\begin{aligned}
\max\;\; & \sum_{i = 1}^{n-1} x_i + y_i \\
s.t. \;\; &  x_i + y_i \leq 1 \quad \forall i\\
& \sum_{i=1}^{n-1} d_i(x_i - y_i) = d_n\\
& 0\leq x_i, y_i \leq 1 \quad \forall i\\
& x_i, y_i \in \mathbb{Z} \quad \forall i.
\end{aligned}
\right.
\end{equation*}
Here, a die value $d_i$ (for die $i$) is added if $x_i = 1$ and subtracted if $y_i = 1$. We are maximizing the number of dice used, which is given by $(x_1 + y_1) + \cdots (x_{n-1}+y_{n-1})$ because the first constraint $x_i + y_i \leq 1$ along with the third and fourth constraints ensure that exactly one of $x_i$ or $y_i$ is $1$ but both may be zero (if a die is not used). The second constraint ensures that the resulting sum equals the value shown on the $n^\text{th}$ die (the dodecahedron).  Using this formulation, the 23,328 distinct Math Dice Jr. rolls (with 6 dice) can be classified in $19.05\,\mathrm{s}$. 

\begin{figure}[htbp]
\centering
\includegraphics[width=0.355\columnwidth]{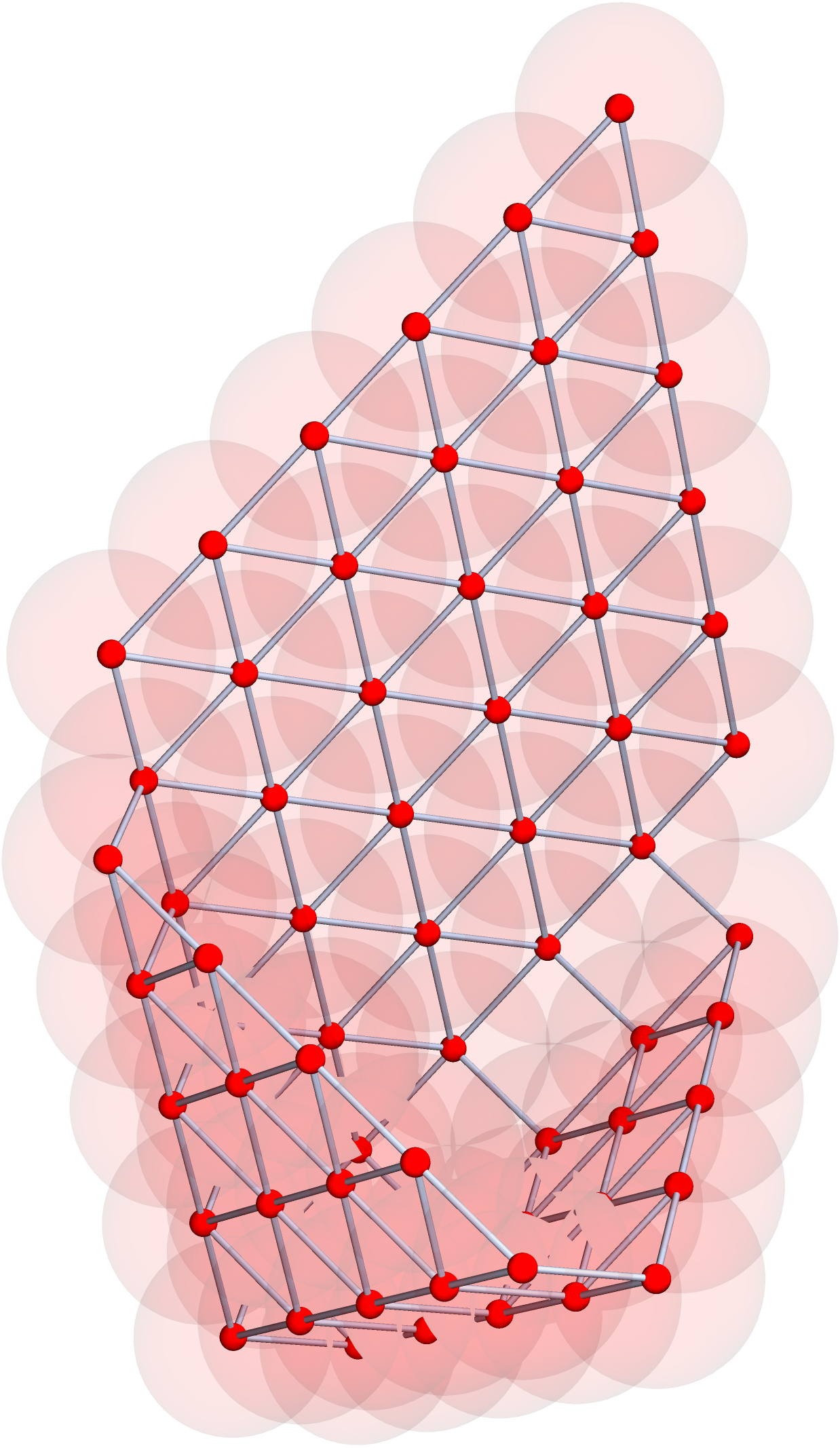} \quad
\includegraphics[width=0.55\columnwidth]{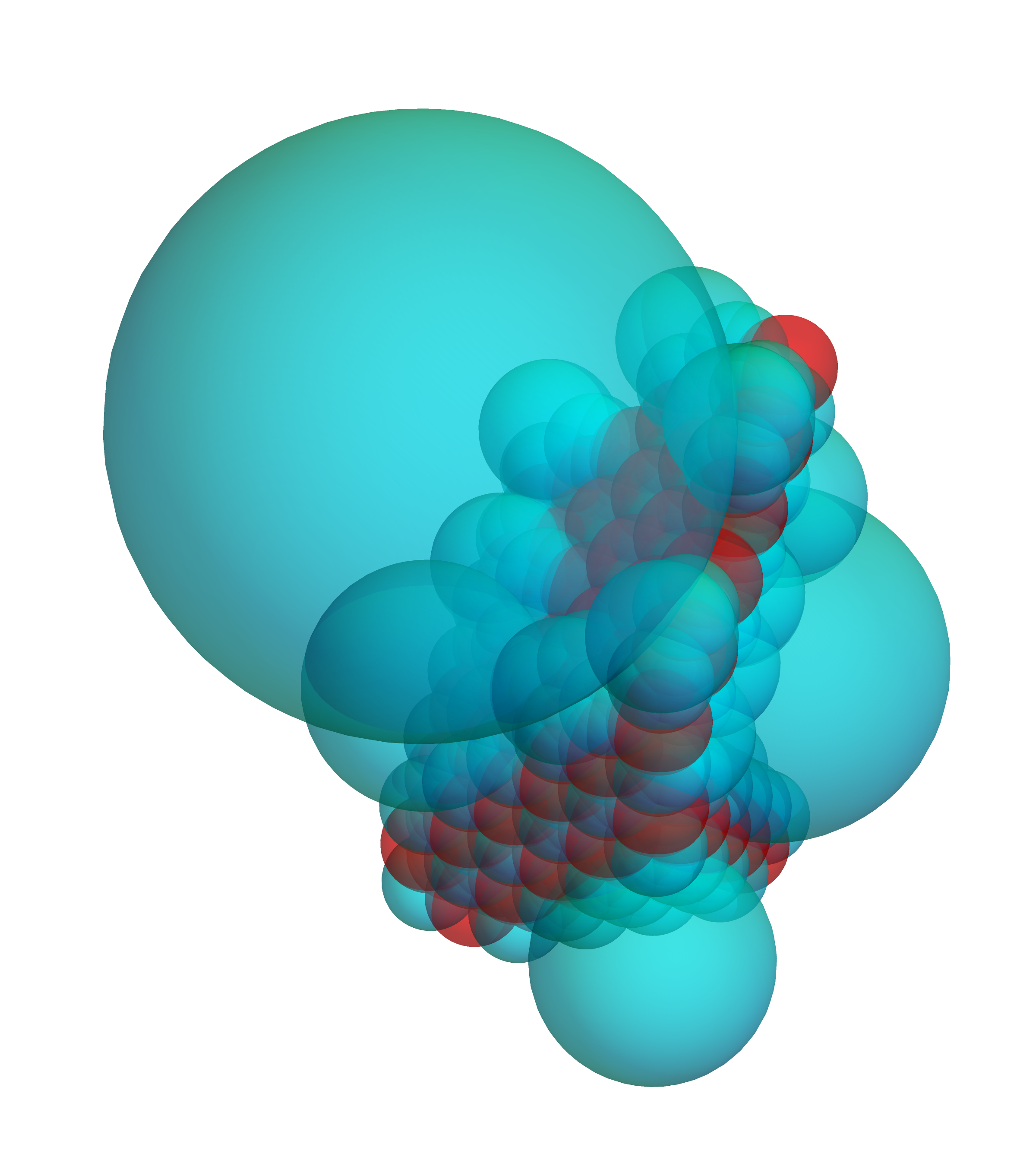}
\caption{(Left) The 1-skeleton $S_1$ and covering for the manifold $\mathcal{M}_1$ on which dice rolls in Math Dice Jr. with 3 dice can yield the value on the dodecahedral dice. (Right) The open cover of both $\mathcal{M}_0$ and $\mathcal{M}_1$ showing the two manifolds wrap around and intersect each other.}
\label{fig:2DiceSurface}
\end{figure}

Using the proposed topological covering algorithm, we can construct simplicial complexes for topological spaces on which the two classes of data lie \footnote{We are being a bit glib here. These spaces are discrete, but we are embedding them in a continuous object.}. In the case when $n=3$, we can visualize the resulting structures. \cref{fig:2DiceSurface} (left) shows that when $n=3$, the topological space containing $X_1$ has a depression in which elements of $X_0$ lie. This causes a void to emerge in the homology of $H_0$ (the simplicial complex corresponding to Class 0). Likewise, the balls defining the open covering of $\mathcal{T}_0$ protrude through $\mathcal{T}_1$ as shown in \cref{fig:2DiceSurface} (right) creating holes. We note that \textit{every} point in this case (and all cases that follow), every element of $X_1$ must be used to create the topological cover.
This leads to the Betti sequence shown in Row 1 of \cref{tab:Topology}. Here, we interpret $\beta_i$ for $i \geq 1$ as the number of $i$-dimensional holes (voids) in the simplicial complex. The number of connected components is $\beta_0$. 

We can compute the proportion of the data used in creating the cover to see the boundary between the two manifolds increases in complexity as the number of dice increase. This is shown in \cref{tab:CoverPropDice}.
\begin{table}[htbp]
\begin{tabular}{|c|c|c|}
\hline
\textbf{\# Dice} & \textbf{Cover 1 Prop.} & \textbf{Cover 0 Prop.}\\
\hline
3 & 1 & 0.355\\
\hline
4 & 1 & 0.615\\
\hline
5 & 1 & 0.853\\
\hline
6 & 1 & 0.957\\
\hline
\end{tabular}
\caption{The proportion of the data used in covers for Class 0 and Class 1 in Math Dice Jr. for varying numbers of dice.}
\label{tab:CoverPropDice}
\end{table}
We note that Class 1 requires all data points to be used in the cover (as is clear from \cref{fig:2DiceSurface}). As the number of dice increases, the number of data points needed to form a covering of Class 0 also increases. This suggests that the structure of the boundary between the manifolds is becoming more complex as the number of dice increase. Because so much of the data set is used to form the topological cover, we did not test the topological classification algorithm on this data set. Instead, we will use the derived topological properties to explore the relationship between learning and topology in various feedforward neural networks.

\begin{figure}[htbp]
\centering
\includegraphics[width=0.8\columnwidth]{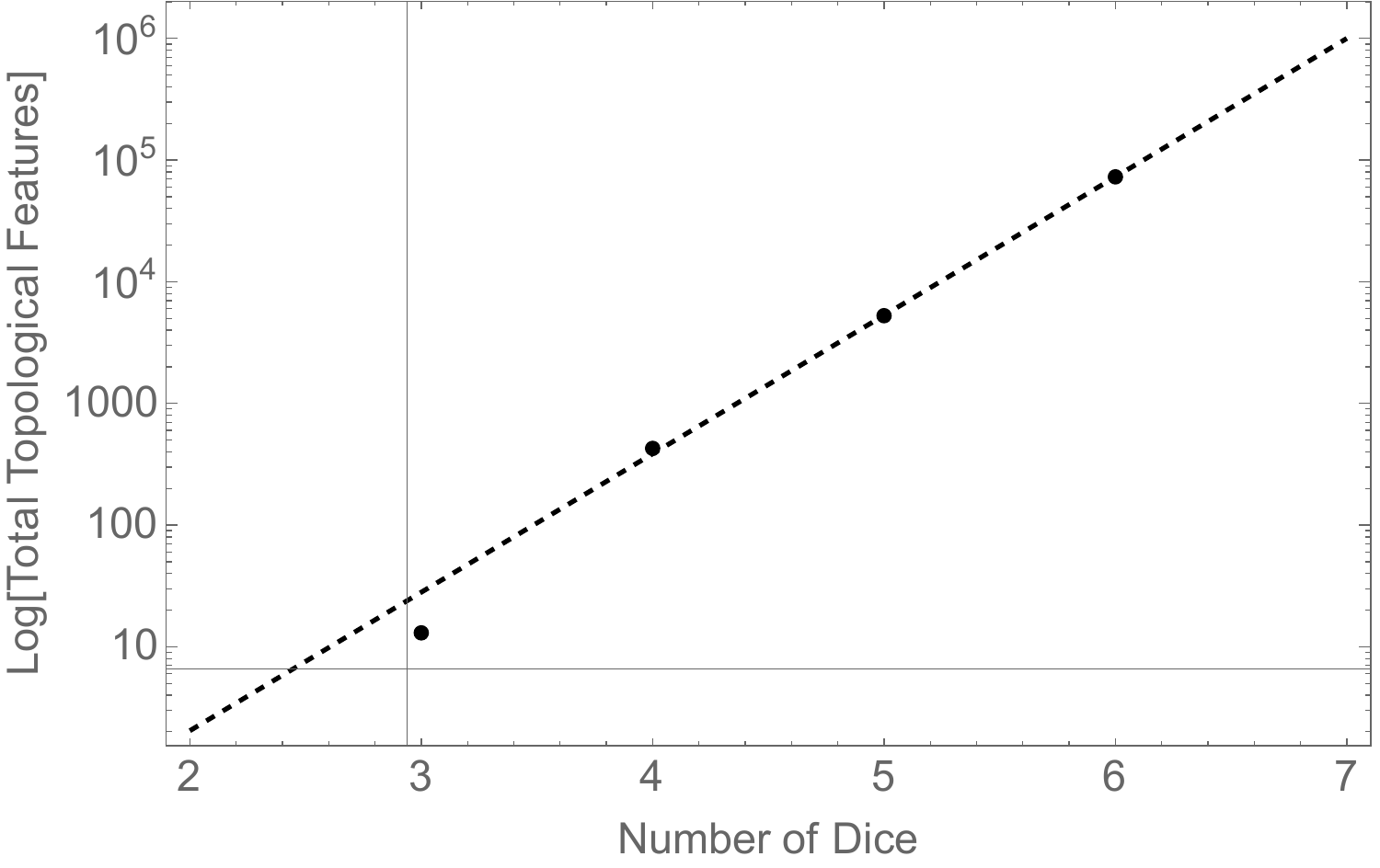}
\caption{The number of topological features in $\mathcal{T}_1$ increases exponentially as a function of the number of dice in Math Dice Jr. suggesting a complex boundary is forming between $\mathcal{T}_0$ and $\mathcal{T}_1$.}
\label{fig:TopFeaturesVsDice}
\end{figure}

\begin{figure}[htbp]
\centering
\includegraphics[width=0.8\columnwidth]{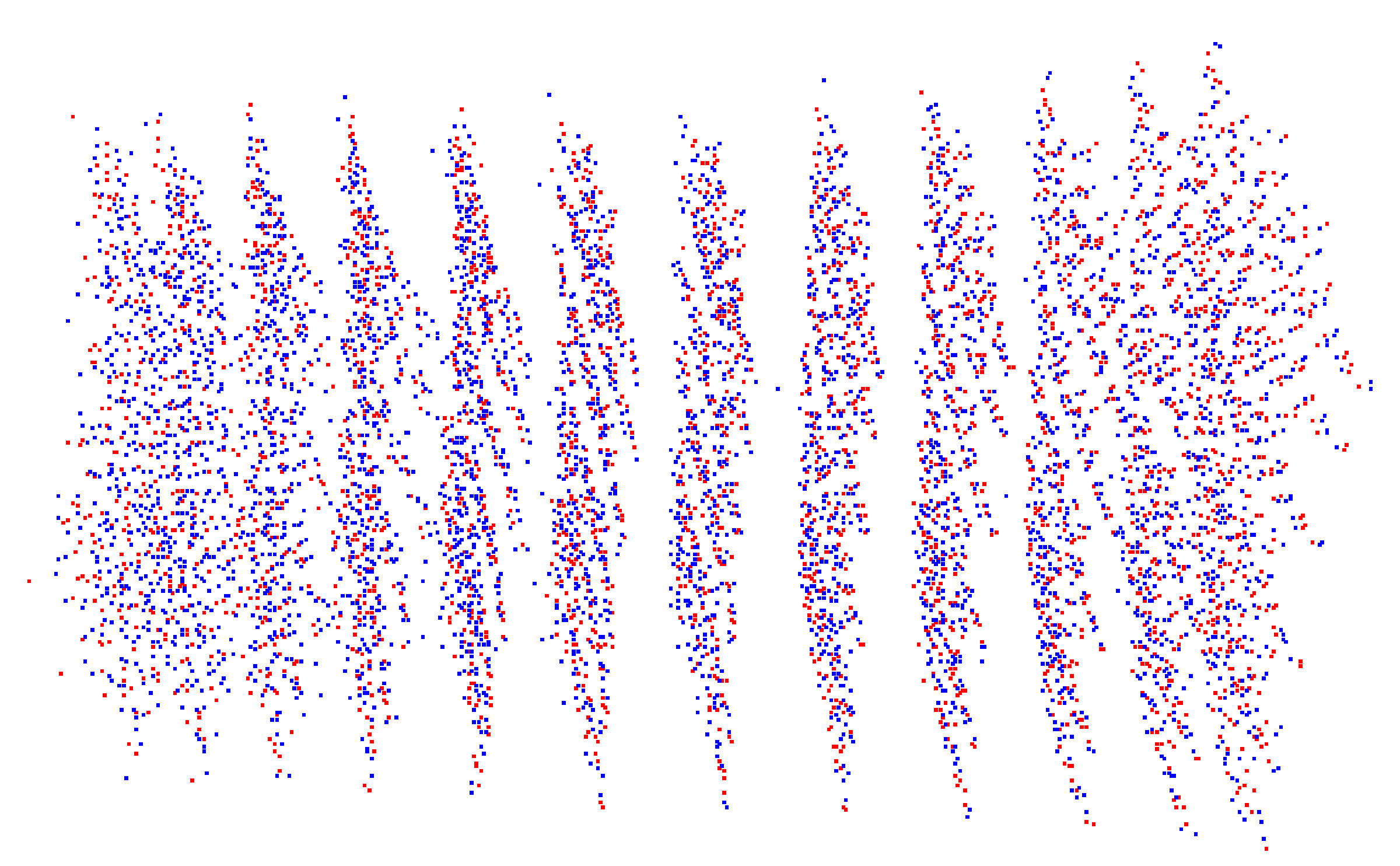}
\caption{Projection of witness vertices of five dice Math Dice Jr. embedded into $\mathbb{R}^2$.}
\label{fig:MathDice4Projection}
\end{figure}

We can quantify the complexity of the boundary explicitly using the Betti numbers of the simplicial complex modeling Class 1. That is, we compute homology for $H_1 = \mathrm{Cl}(C_1^*)$ for $n=3,\dots,6$. Betti numbers are shown in subsequent rows of \cref{tab:Topology}.
\begin{table}[htbp]
\begin{tabular}{|c|c|c|c|c|c|c|}
\hline
{\rule[0.6mm]{4em}{.5pt}} &  \multicolumn{6}{c|}{\textbf{Betti Numbers}} \\
\hline
\textbf{\# Dice} & $\beta_0$ & $\beta_1$ & $\beta_2$ & $\beta_3$ & $\beta_4$ & $\beta_5$ \\
\hline
3 & 1 & 12 & 0 & 0 & 0 & 0 \\
\hline
4 & 1 & 0 & 357 & 69 & 0 & 0 \\
\hline
5 & 1 & 0 & 725 & 4,522 & 12 & 0 \\
\hline
6 & 1 & 0 & 411 & 72,093 & 250 & 75\\
\hline
 \end{tabular}
\caption{The Betti numbers counting the numbers of components, holes and voids in the manifold $\mathcal{M}_1$ as estimated by the homology of $\mathrm{Cl}(S_1)$ shows more complex structure as $n$ increases.}
 \label{tab:Topology}
\end{table}
Define the total number of topological features as,
\begin{equation*}
T = \sum_{i=0}^\infty \beta_i.
\end{equation*}
As the number of dice increases, the data suggest  that the total number of topological features increases exponentially (see \cref{fig:TopFeaturesVsDice}).
We conjecture that as the number of dice increases, the structure of the boundary between $\mathcal{T}_0$ and $\mathcal{T}_1$ becomes more complex, as evidenced by the exponentially increasing number of topological features in $\mathcal{T}_1$ as measured by the homology of $H_1$ as well as the proportion of the data needed to construct a covering of the two data sets. We illustrate the complexity of the boundary for the five dice case by constructing a joint skeleton as before. The image (\cref{fig:MathDice4Projection}), in which edges are suppressed for clarity, shows the two classes of data are thoroughly mixed  as compared to (e.g.) \cref{fig:HEPMASSCover} which is easily separable. 

An increasingly complex boundary suggests that a more complex (i.e. deeper and/or wider) neural network structure is required to learn such a boundary in order to classify a sample. \cref{tab:AccuracyVsNet} shows evidence to support this hypothesis. 
\begin{table}[htbp]
\begin{tabular}{|c|c|c|c|}
\hline
\textbf{Dice} & \textbf{DNN Struct.} & \textbf{Mean Acc.} & \textbf{(Min, Max)} \\
\hline
3 & $(16,4,2)$ & 0.912 & $(0.787,1.)$ \\
\hline
 3 & $(8,4,2)$ & 0.881 & $(0.787,1.)$ \\
 \hline
 4 & $(32,4,2)$ & 0.954 & $(0.922,0.988)$ \\
 \hline
 4 & $(16,4,2)$ & 0.816 & $(0.922,0.988)$ \\
 \hline
 5 & $(128,16,2)$ & 0.828 & $(0.798,0.867)$ \\
 \hline
 5 & $(256,16,2)$ & 0.832 & $(0.798,0.867)$ \\
 \hline
 6 & $(1024,256,64,16,16,2)$ & 0.553 & $(0.529,0.592)$ \\
 \hline
 6 & $(2048,256,64,16,16,2)$ & 0.547 & $(0.529,0.592)$ \\
 \hline
\end{tabular}\\
\textbf{Funnel Design}\\

\begin{tabular}{|c|c|c|c|}
\hline
\textbf{Dice} & \textbf{DNN Struct.} & \textbf{Mean Acc.} & \textbf{(Min, Max)} \\
\hline
3 & $(16,16,2)$ & 1. & $(1.,1.)$ \\
\hline
 3 & $(8,8,2)$ & 0.942 & $(1.,1.)$ \\
 \hline
 4 & $(32,32,2)$ & 0.989 & $(0.979,1.)$ \\
 \hline
 4 & $(16,16,2)$ & 0.865 & $(0.979,1.)$ \\
 \hline
 5 & $(128,128,2)$ & 0.935 & $(0.882,0.961)$ \\
 \hline
 5 & $(256,256,2)$ & 0.934 & $(0.882,0.961)$ \\
 \hline
 6 & $(256,256,256,2)$ & 0.531 & $(0.538,0.786)$ \\
 \hline
 6 & $(1024,1024,1024,2)$ & 0.656 & $(0.538,0.786)$ \\
 \hline
 6 & $(2048,2048,2048,2)$ & 0.536 & $(0.522,0.55)$ \\
 \hline
\end{tabular}\\
\textbf{Flat Design}
\caption{(Top) Learning results for various funnel design DNN's on the Math Dice Jr. problem. (Bottom) Learning results for various flat DNN's on the Math Dice Jr. Problem. In both cases, even when the complexity of the neural network structure required grows super-exponentially, a 5-layer DNN with over $10^9$ connections cannot learn the boundary structure between $\mathcal{C}_1$ and $\mathcal{C}_0$ in 6 dice Math Dice Jr.}
\label{tab:AccuracyVsNet}
\end{table}
Accuracy results are computed over 20 random train/test splits. We evaluated both flat and funnel architectures in the DNN's. In this experiment, the neural network complexity (as measured in number of edges) grows super-exponentially (as compared to the exponential growth in topological features). The most complex DNN used for the 6 dice case has over $10^9$ connections. However, this network is incapable of separating $X_1$ from $X_0$. 

To further understand why a large DNN fails on Math Dice Jr. with 6 dice, we analyze the behavior of DNN's that can learn Math Dice Jr. with 5 dice to determine how each layer in the network changed the topological structure of the data. Formally, we think of layer $i$ as a function $f_i:\mathbb{R}^{n_{i-1}} \to \mathbb{R}^{n_i}$. If we exclude the final classification step, the neural network is simply a function $f:\mathbb{R}^N \to \mathbb{R}^2$ such that $f = f_{L}\circ \cdots f_{1}$. If $f$ is a homeomorphism, then $f$ cannot change the homology of the underlying manifolds. However, since we are interested in homological features up to the level of resolution of (e.g.) Class $0$ within Class $1$, we expect to see a topological simplification taking place at each level. The data in \cref{tab:BettiNumbersLayers} support this hypothesis. 
\begin{table}[htbp]
\begin{tabular}{|c|c|c|c|c|c|c|}
\hline
{\rule[0.6mm]{4em}{.5pt}} &  \multicolumn{6}{c|}{\textbf{Betti Numbers}}\\
\hline
\textbf{Layer} & $\bar{\beta }_0$ & $\bar{\beta }_1$ & $\bar{\beta }_2$ & $\bar{\beta }_3$ &
   $\bar{\beta }_4$ & $\bar{\beta }_5$ \\
   \hline
 0 (Input) & 1 & 0 & 725 & 4,522 & 12 & 0 \\
 \hline
1 & 1 & 0 & 0 & 570. & 303. & 0. \\
\hline
 2 & 20.2 & 12.3 & 0.9 & 0.1 & 0 & 0 \\
 \hline
 3 & 9.45 & 2.4 & 0 & 0 & 0 & 0 \\
 \hline

\end{tabular}
\caption{The topological complexity of the boundary between $\mathcal{M}_0$ and $\mathcal{M}_1$ is further simplified and refined at each layer of a neural network that successfully classifies Math Dice Jr.}
\label{tab:BettiNumbersLayers}
\end{table}
The data were generated using the complete 5 dice Math Dice Jr. data set and training a neural network with structure $(256,16,2)$. The classifier layer was then removed and the original 5 dice data set was transformed (by $f_{L-1}\circ \cdots f_{1}$). Homological information on the clique complex for this lower dimensional data set was then computed. This process was repeated for hidden layers two and one. We repeated this in 20 replications (to average out effects from stochastic gradient descent). \cref{tab:BettiNumbersLayers} shows mean Betti numbers (indicating topological structure) for these 20 runs. 
\begin{figure}[htbp]
\centering
\includegraphics[width=0.8\columnwidth]{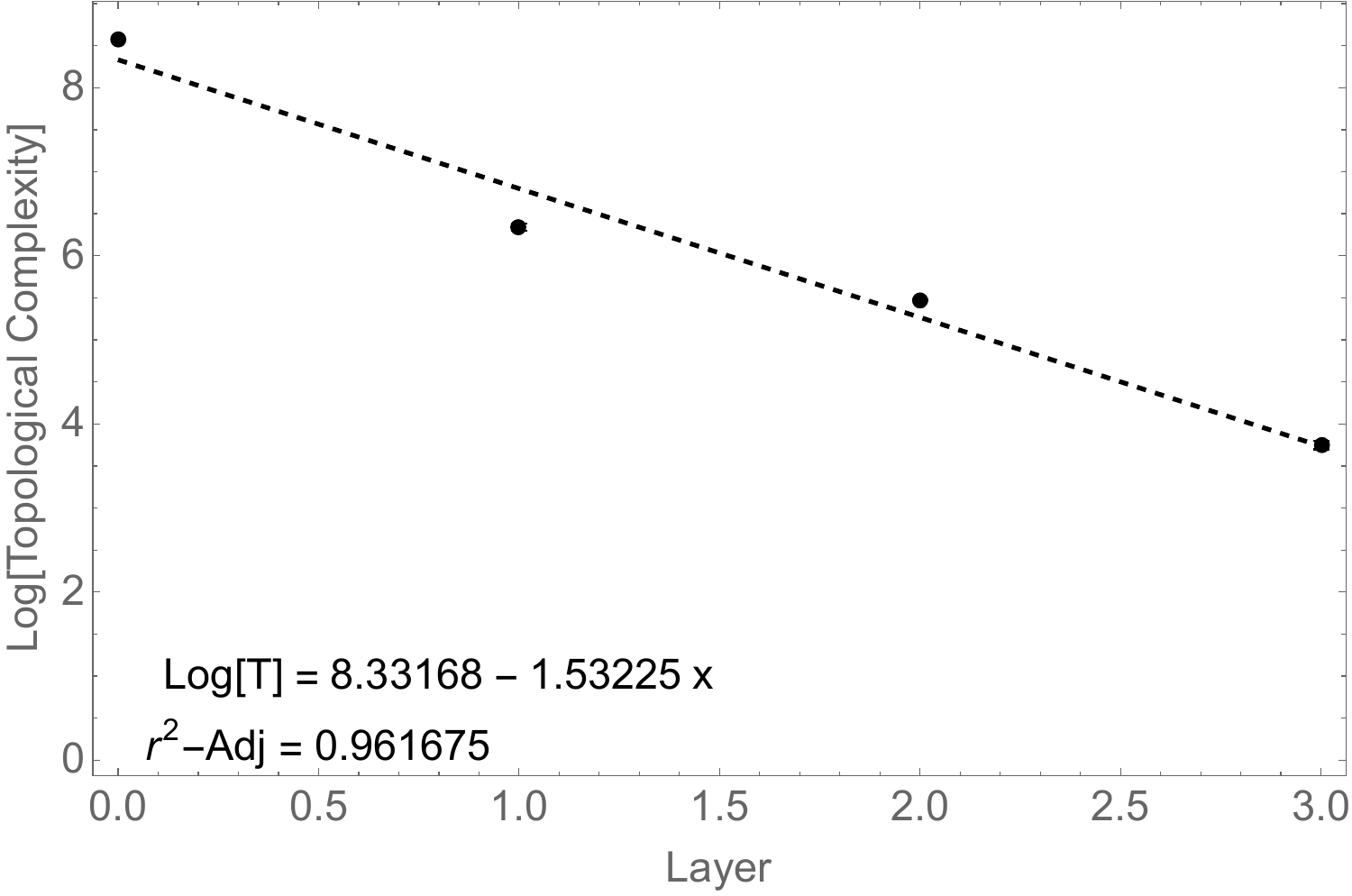}
\caption{The topological complexity of the transformed data decreases exponentially with each layer, ultimately making classification trivial.}
\label{fig:MathDice4TopologyEffect}
\end{figure}
As shown in \cref{fig:MathDice4TopologyEffect}, the number of topological features in the data decreases exponentially in each layer. Note this is similar to the results identified by Naitzat, Zhitnikov and Lim \cite{NZL20} in their independent study. This supports the hypothesis that each layer is simplifying the boundary structure between the two classes of data. We compare this to the topological complexity of the data output by the last layer of a neural network with structure $(2048,256,64,16,2)$ acting on 6 dice Math Dice Jr. data, which we know from \cref{tab:AccuracyVsNet} cannot successfully identify the boundary structure between $\mathcal{T}_0$ and $\mathcal{T}_1$. In an example run, the data produced by the final layer is disconnected into several thousand components, showing that the neural network has mapped the two classes onto each other, explaining the confusion and low accuracy in the last row of \cref{tab:AccuracyVsNet}. We leave a general investigation of why the flat and funnel architectures failed and how this relates to the topological complexity as future research.

By the universal approximation theorem \cite{C89,C92} there is a feedforward neural network with sigmoid activations functions that can separate $X_0$ and $X_1$ in the 6 dice Math Dice Jr. case. However, finding the simplest such neural network structure is clearly non-trivial, consistent with the no free lunch theorem \cite{W96,WM97}. Nevertheless, we can use geometric information to construct a neural network architecture that for this problem. 

Let $\mathcal{W} = \{-1,1\}^5 \subset \mathbb{R}^5$. Let $\mathcal{R}\subset\mathbb{R}^6$ be the set of possible rolls in 6 dice Math Dice Jr. For $\mathbf{w} \in \mathcal{W}$ and $\mathbf{x}\in\mathcal{R}$, let:
\begin{equation*}
\varphi(\mathbf{x};\mathbf{w}) = \sum_{i=1}^5 w_i x_i - x_6.
\end{equation*}
Let:
\begin{equation*}
\tilde{\delta}(x) = 
\begin{cases} 1 & x = 0\\
0 & \text{otherwise}.
\end{cases}
\end{equation*}
Choose an order so that $\mathcal{W} = \{\mathbf{w}_1,\dots,\mathbf{w}_{32}\}$. Then the function, $\bm{F}:\mathbb{R}^6 \to \mathbb{R}^{32}$ defined by,
\begin{equation*}
\mathbf{F}_i(\mathbf{x}) = \tilde{\delta}\left[\varphi(\mathbf{x};\mathbf{w}_i)\right],
\end{equation*}
maps each roll to a vertex on the unit hypercube $\mathbb{R}^{32}$. For a roll $\mathbf{x}$, all five hexahedral dice can be used to obtain the number on the dodecahedral die if and only if there is an $i$ so that $\mathbf{F}_i(\mathbf{x}) = \mathbf{0}$ (the vector of all zeros). Any linear separator that separates the vertex $\mathbf{0}$ from the other vertices of the unit hypercube in $\mathbb{R}^{32}$ will correctly classify this point. This is the explicit mapping in Cover's theorem \cite{C65}.

\begin{figure}[htbp]
\centering
\includegraphics[width=0.8\columnwidth]{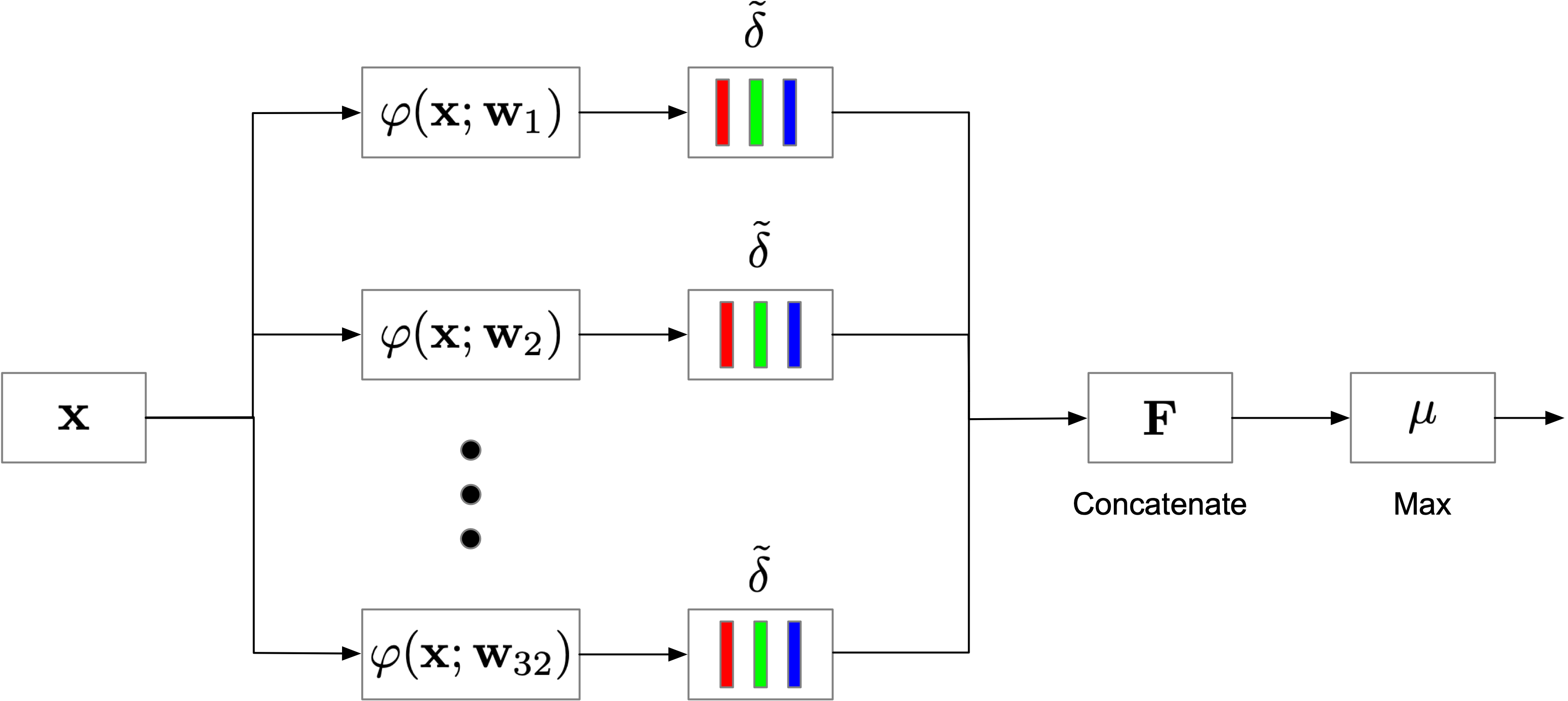}
\caption{A custom feed forward neural network architecture that will correctly classify all rolls in 6 dice Math Dice Jr. This architecture can be easily generalized to arbitrary Math Dice Jr. games.}
\label{fig:CustomNN}
\end{figure}

For simplicity, let $\sigma:\{0,1\}^{32} \to \{0,1\}$ so that:
\begin{equation*}
\mu(\mathbf{y}) = \max_{i} \mathbf{y}_i.
\end{equation*}
Then the function that classifies any roll $\mathbf{x} \in \mathbb{R}$ is given by:
\begin{equation*}
C(\mathbf{x}) = \mu[\mathbf{F}(\mathbf{x})] = \mu\{\tilde{\delta}\left[\varphi(\mathbf{x};\mathbf{w}_i)\right]\}.
\end{equation*}
This can be encoded in the feedforward neural network architecture shown in \cref{fig:CustomNN}.
It is straightforward to generalize this architecture for an arbitrary Math Dice Jr. game by scaling $\mathcal{W}$. Note, this scales exponentially, as is to be expected since Math Dice Jr. is $\mathrm{NP}$-hard for arbitrary numbers of dice.

We can formally analyze the topological structure of the data produced by the layers of this neural network. We note that using the mapping $\mathbf{F}(\mathbf{x})$ to transform the data into thirty-two dimensional points creates a 1-skeleton that is the complete graph on the 49 open spheres that can be used to cover $\mathbf{F}(\mathcal{R})$. Thus, all voids are closed by this mapping and the data fully separates into two manifolds that are contractible to a single point.  

\begin{figure}[htbp]
\centering
\includegraphics[width=0.45\columnwidth]{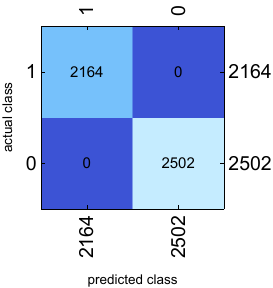}\quad
\includegraphics[width=0.45\columnwidth]{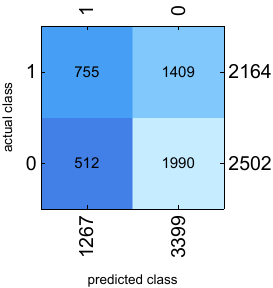}
\caption{(Left) Confusion matrix for the pre-specified feedforward neural network with weights given by $\mathcal{W}$. (Right) Confusion matrix for the neural network with structure shown in \cref{fig:CustomNN}, but with weights trained using stochastic gradient descent.}
\label{fig:PerfectMathDice}
\end{figure}

For 6 dice Math Dice Jr., the function $\hat{\delta}$ can be approximated using,
\begin{equation*}
\tilde{\delta}(x) \approx 1 - \tanh\left(10 |x|\right),
\end{equation*}
since we are using integer data. Using this, a neural network was constructed with perfect accuracy. We then did an experiment, where we removed the weights specified in $\mathcal{W}$ and allowed them to be trained using stochastic gradient descent on a random 80/20 train/test split. The resulting accuracy was $58.5\pm0.7\%$. This is similar to the result shown in \cref{tab:AccuracyVsNet} and shows the challenge in learning to play Math Dice Jr. with 6 dice, even when the structure of a successful feedforward neural network is known. Confusion matrices for the two cases are shown in \cref{fig:PerfectMathDice}.

\section{Conclusion and Future Directions}\label{sec:Conclusions}
In this paper, we developed a novel approach to topological data analysis on multi-class data that allows us to build simplicial complex representations of the data sets that faithfully represent the features of the data at the scale of the classes. This allowed us to develop a topological classifier that is competitive with random forest and DNN classifiers. This was validated on multiple publicly available data sets. The topological information that can be extracted from this approach also provides additional insights into data structure. Additional analysis on data generated from the game Math Dice Jr. and a custom two-dimensional data set with special topological features supports a hypothesis that topological complexity in the boundary between classes can lead to learning failure in DNN's using stochastic gradient descent. This suggests further research into this hypothesis is warranted, as it may provide insights into scenarios where AI/ML solutions will fail in complex physics problems \cite{S23}. We also note, that while the approach discussed in this paper performs well on the chosen data sets, the topological approach does not scale (in time) as well as DNN methods. In particular, it took well over 24 hours to generate a topological cover for the HEPMASS data set, while it took only minutes to train the DNN that successfully classified the data set. On smaller data sets, the computation time was much more comparable. As a consequence, we advocate this analysis method only in cases where deep inspection of the topological structure of the data is warranted. 

The results presented in this paper suggest several future directions of research. Additional experimentation with data sets with topologically complex boundaries and the resulting learning problems in DNN's seems worthy of investigation. In particular, exploring the impact on the loss function and the process of stochastic gradient descent in these scenarios may yield insight into the nature of learning failure. Additionally, an investigation into the impact the choice of metric has on this approach would provide additional insight. For example, for data sets that contain both categorical and continuous data, the ability to design custom metrics and apply the proposed topological analysis might yield insights about the structure of the data itself. Finally, finding novel approaches to computing Betti numbers for large simplicial complexes may provide additional insights into scenarios where DNN's function well. 

\section*{Data Availability} 
Mathematica notebooks for analyzing all data sets, excluding HEPMASS and the Scripps Institute acoustic data, are available as supplemental information \footnote{See Supplemental Material at [URL will be inserted by publisher] for a complete set of Mathematica notebooks.}. For the C++ code used in analyzing the large-scale data, please contact the authors. 
Example code is also available on the Wolfram Community site \cite{GKA23}.

\appendix
\section{Alternate Covering Algorithm}\label{app:AltAlg}
When it is computationally difficult to construct $\vec{G}(C_i)$, the following algorithm can be used to construct a smaller sub-cover $C_i^*$ instead.
\begin{algorithm}[H]
  \caption{Approximate Minimal Sub-Cover - 2}
  \label{alg:AMSCr}
   \begin{algorithmic}[1]
   \State Sort the elements of $C_i$ by order of decreasing radius.
   \ForAll{$(x_{i_j},r_{i_j}) \in C_i$}
   \If{there does not exist $(x_{i_k}, r_{i_k}) \in C_i^*$ that covers $(x_{i_j}, r_{i_j})$}
   \State Add $(x_{i_j}, r_{i_j})$ to $C_i^*$.
   \EndIf
   \EndFor
   \end{algorithmic}
\end{algorithm}
This algorithm is a variant of the canopy clustering algorithm \cite{KIYP+14}, but with class information defining the distances to be used. 

\bibliography{TopologyPaper}

\end{document}